\documentclass[10pt,twocolumn,letterpaper]{article}

\usepackage[accsupp]{axessibility}  

\usepackage{cuted}         
\usepackage{caption}       

\usepackage[pagenumbers]{wacv} 

\usepackage{algorithmic}
\usepackage{caption}

\usepackage{algorithm}
\usepackage{graphicx}

\usepackage{tcolorbox}
\tcbuselibrary{listingsutf8}
\usepackage[dvipsnames]{xcolor}

\definecolor{learnbg}{RGB}{230,240,255}   
\definecolor{genbg}{RGB}{255,230,230} 

\newtcolorbox{genblock}{
  colback=genbg, colframe=genbg,
  boxrule=0pt, arc=2pt, left=2pt, right=2pt, top=2pt, bottom=2pt
}
\newtcolorbox{learnblock}{
  colback=learnbg, colframe=learnbg,
  boxrule=0pt, arc=2pt, left=2pt, right=2pt, top=2pt, bottom=2pt
}

\newcommand{\RNum}[1]{\uppercase\expandafter{\romannumeral #1\relax}}

\definecolor{wacvblue}{rgb}{0.21,0.49,0.74}
\usepackage[pagebackref,breaklinks,colorlinks,allcolors=wacvblue]{hyperref}

\def\wacvPaperID{1326} 
\def\confName{WACV}
\def\confYear{2026}

\title{End-to-End Fine-Tuning of 3D Texture Generation using Differentiable Rewards}

\author{\parbox{\textwidth}{\centering
AmirHossein Zamani$^{1,2}\;$ \thanks{Correspondence to: {\tt\small amirhossein.zamani@mila.quebec}} \hspace{-7pt}
\qquad
Tianhao Xie$^{2}$\hspace{-15pt}
\qquad
Amir G. Aghdam$^{2}$\hspace{-15pt}
\qquad
Tiberiu Popa$^{2}$\hspace{-15pt}
\qquad
Eugene Belilovsky $^{1,2}$ \hspace{-15pt}
} \vspace{5pt}\\
 $^1$ Mila -- Quebec AI Institute; $^2$ Concordia University, Montreal, QC, Canada  \\
}

\begin{document}
\maketitle

\begin{strip}
  \centering
  \vspace{-55pt}
  \includegraphics[scale=0.22]{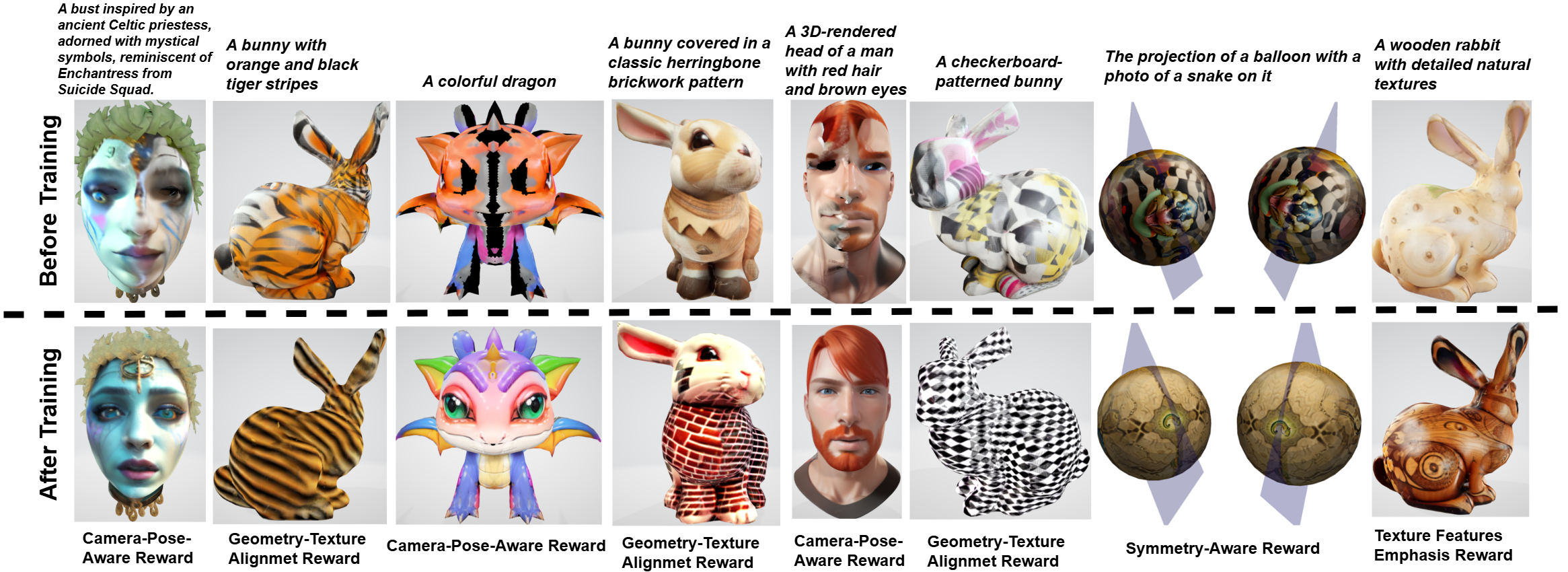}
  \captionof{figure}{%
    Texture images generated before and after reward fine‑tuning on different 3D mesh objects
    using different rewards. Each column shows the text prompt at the top and the corresponding
    reward function used for texture generation at the bottom. Texture images produced by our method
    consistently outperform the pre‑fine‑tuning baseline (InTeX \cite{InTex}) across all rewards and experiments.%
  }
  \label{fig:teaser}
\end{strip}

\begin{abstract}   
    While recent 3D generative models can produce high-quality texture images, they often fail to capture human preferences or meet task-specific requirements. Moreover, a core challenge in the 3D texture generation domain is that most existing approaches rely on repeated calls to 2D text-to-image generative models, which lack an inherent understanding of the 3D structure of the input 3D mesh object. To alleviate these issues, we propose an end-to-end differentiable, reinforcement-learning-free framework that embeds human feedback, expressed as differentiable reward functions, directly into the 3D texture synthesis pipeline. By back-propagating preference signals through both geometric and appearance modules of the proposed framework, our method generates textures that respect the 3D geometry structure and align with desired criteria. To demonstrate its versatility, we introduce three novel geometry-aware reward functions, which offer a more controllable and interpretable pathway for creating high-quality 3D content from natural language. By conducting qualitative, quantitative, and user-preference evaluations against state-of-the-art methods, we demonstrate that our proposed strategy consistently outperforms existing approaches. Our implementation code is publicly available at: \href{https://github.com/AHHHZ975/Differentiable-Texture-Learning}{https://github.com/AHHHZ975/Differentiable-Texture-Learning}
\end{abstract}

\vspace{-15pt}
\section{Introduction}
\label{sec:introduction}
{
    \vspace{-5pt}
    \begin{figure*}[h]
        \vskip -0.05in
        \begin{center}
        \centerline{\includegraphics[scale=0.235]{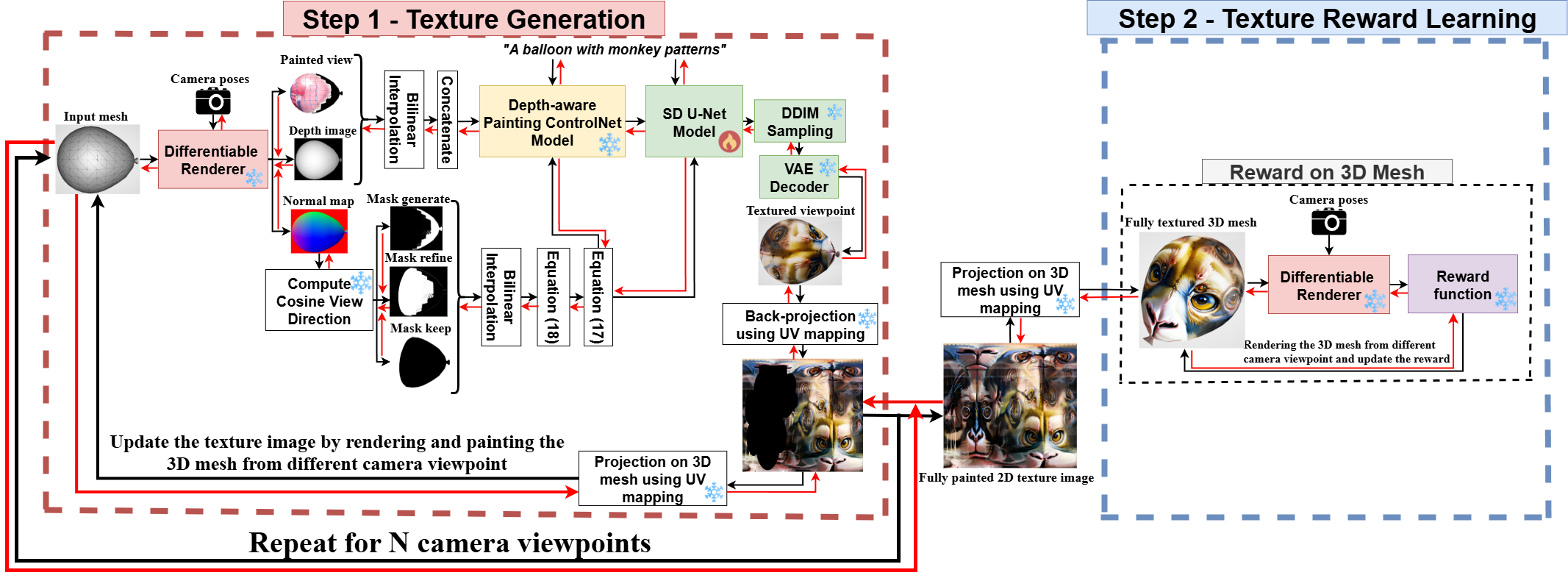}}
        \caption{An overview of the proposed training process, consisting of two main stages: (i) \textit{\textcolor{red}{texture generation}} (\cref{TextureGenerationStep}), where a latent diffusion model generates high-quality images from textual prompts. Combined with differentiable rendering and 3D vision techniques, this step produces realistic textures for 3D objects. (ii) \textit{\textcolor{blue}{texture reward learning}} (\cref{TextureEnahncementStep}), where an end-to-end differentiable pipeline fine-tunes the pre-trained text-to-image diffusion model by maximizing a differentiable reward function $r$. Gradients are back-propagated through the entire 3D generative pipeline, making the process inherently geometry-aware. To demonstrate the method’s effectiveness in producing textures aligned with 3D geometry, we introduce five novel geometry-aware reward functions, detailed in \cref{GeometryAwareRewardDesign} and \cref{GeometryGuidedTextureColorizationReward}.}
        \label{TrainingOverview}
        \end{center}
        \vspace{-30pt}
    \end{figure*}
    
    While large-scale generative computer vision models learn broad knowledge and some reasoning skills to generate images \cite{DDPO, DPOK}, videos \cite{VideoFromText}, and 3D objects \cite{DreamFusion, Magic3D}, achieving precise control of their behavior is difficult due to the completely unsupervised nature of their training. Among these generative models, 3D generative ones often rely on adapting text-to-image models at inference \cite{DreamFusion, Magic3D, Paint3D}, but the resulting 3D content may not align with human preferences or task-specific needs. This challenge highlights the need for a preference learning frameworks adapted to 3D content creation. One general solution proposed mainly in the large language models' literature \cite{InstructGPT, DeepRLHF, RAFT, DPO, RLEnahnced_LLMs} is to incorporate human or automated feedback as a supervisory signal to guide the generative models toward desired behavior. This is usually done by leveraging reinforcement learning from human feedback (RLHF) \cite{DeepRLHF}. Prior work has shown that RLHF can significantly improve results in some domains, such as text generation \cite{InstructGPT, RAFT} and image synthesis \cite{DDPO, DPOK, Draft}. However, to the best of our knowledge, DreamReward \cite{DreamReward} is the only prior work that incorporates human feedback into 3D generative model training. It applies differentiable rewards to optimize NeRF-based geometry editing, without training the generative model. In contrast, we directly fine-tune the diffusion model for texture editing within a 3D generation pipeline, enabling precise control aligned with user preferences. Moreover, unlike DreamReward’s aesthetic rewards that ignore geometry, our geometry-aware rewards ensure textures are both perceptually rich and structurally aligned with the geometry of the input 3D mesh. Despite recent advances, a core challenge in 3D texture generation remains: most existing approaches rely on repeated calls to 2D text-to-image models, which lack an inherent understanding of 3D structure. Therefore, applying preference learning directly to these 2D components often leads to solutions that neglect crucial 3D constraints. To address this, we propose an end-to-end differentiable preference learning framework that back-propagates differentiable rewards through the entire 3D generative pipeline, making the process inherently geometry-aware. By directly fine-tuning the 3D texture generation model through differentiable geometry-aware rewards, our framework supports interactive and user-aligned texture editing, producing results that are both visually compelling and geometrically coherent. The main contributions of this study are:    
    \begin{itemize}
        \item We propose a method to fine-tune a 2D text-to-image foundational model within a 3D generative pipeline in an end-to-end fashion using 3D geometry-aware rewards.
        \item We propose several novel geometry-aware rewards that can be used in conjunction with our end-to-end differentiable framework to improve texture generation on challenging 3D shapes and realistic user requirements.Empirically validating their performance.
    \end{itemize}    
}

\section{Related Work}
\label{sec:related_work}
{    

   \paragraph{Texture Generation.}
        Modern 3D texture generation methods leverage text-to-image or image-to-image diffusion models \cite{Imagen, DiffusionModels, StableDiffusionModel} to produce high-quality, realistic textures. These techniques broadly fall into two categories:
    
        \noindent \textit{(i) Texturing via score distillation sampling (SDS).}
    These methods treat 3D shape parameters as learnable and uses SDS as supervision. DreamFusion \cite{DreamFusion} introduced this by optimizing a NeRF \cite{NeRF} whose views are scored by a pre-trained diffusion model. Latent-NeRF \cite{Latent-NeRF} (LatentPaint) extends this by directly optimizing UV texture pixels, and Magic3D \cite{Magic3D} accelerates the process via a two-stage coarse-to-fine pipeline. While these methods yield compelling textures, they remain computationally expensive (1–2 hours per asset), often noisy, and offer limited control due to entangled geometry and appearance. \noindent \textit{(ii) Texturing via direct use of image-generation diffusion models.}
        Alternatively, methods like TEXTure \cite{TEXTure}, Text2Tex \cite{Text2Tex}, and InTex \cite{InTex} assume fixed geometry and generate textures by rendering multi-view images, applying a depth-aware diffusion model, and projecting results back to the UV atlas. This decoupling enables faster runtime and easier editing but lacks a deep understanding of 3D structure, often misaligning textures with human intent or task needs. Our work builds on this paradigm by introducing a geometry-aware fine-tuning stage that aligns textures with human preferences and task-specific objectives, improving coherence and control without the overhead of joint optimization.
    }

    \paragraph{Human Preference Learning.} Recently popularized in the large language model community \cite{InstructGPT, DeepRLHF, RAFT, DPO}, this area leverages human or automated feedback as supervision to steer generative models toward desired behaviors. Typically implemented via reinforcement learning from human feedback (RLHF) \cite{DeepRLHF}, this approach has significantly advanced text generation \cite{InstructGPT, RAFT} and image synthesis \cite{DDPO, DPOK}. In 3D generation, however, its application remains limited. DreamReward \cite{DreamReward} learns NeRF-based shape and appearance using differentiable rewards combined with SDS loss, but due to the entangled nature of NeRF representations, it offers limited control over texture. DeepMesh \cite{DeepMesh}, by contrast, applies direct preference optimization (DPO) \cite{DPO} to mesh geometry alone, neglecting texture modeling entirely. Our approach departs from both. We focus on explicit mesh texturing, not implicit NeRF reconstruction, and fine-tune the diffusion model within a 3D texture-generation pipeline. By designing geometry-aware reward functions that align texture gradients with surface curvature, our method enables precise, user-driven control over texture appearance. The resulting textures are perceptually rich, structurally coherent, and aligned with human preferences or task-specific objectives via differentiable rewards.

    \vspace{-5pt}

\section{Methodology}
\label{sec:methodology}
{   
    \vspace{-5pt}
    We propose a simple, reinforcement-learning-free, end-to-end differentiable framework that can fine-tune a 3D texture generation model to learn any human or task-specific preferences that can be represented as differentiable rewards. The training consists of two stages: (i) \textit{texture generation}, where a diffusion model \cite{StableDiffusionModel} combined with 3D vision techniques like differentiable rendering \cite{NvidiffRast} produces realistic textures from text prompts (\cref{TextureGenerationStep}); and (ii) \textit{texture reward learning}, where the diffusion model is fine-tuned via a differentiable pipeline that maximizes a reward function $r$, enabling gradients to flow through the full 3D generation process and making it geometry-aware (\cref{TextureEnahncementStep}). To demonstrate the method's effectiveness, we introduce five novel geometry-aware differentiable reward functions, each detailed in \cref{GeometryAwareRewardDesign} and \cref{GeometryGuidedTextureColorizationReward}. An overview of the training procedure of the the proposed end-to-end differentiable framework is shown in \cref{TrainingOverview} and \cref{alg:fine-tuning}. The framework supports any form of human preference that can be formulated as a differentiable feedback signal either in the 2D texture space or directly on the 3D surface. This flexibility makes the approach broadly applicable and extensible to a wide range of 3D tasks beyond texture generation.

    \subsection{Texture Generation}
    \label{TextureGenerationStep}
    {    
        Our depth-aware text-to-texture pipeline builds on the methods established in \cite{TEXTure} and \cite{InTex}. Using a differentiable renderer \cite{NvidiffRast}, we render the object from multiple viewpoints to extract inpainted images, depth maps, normal maps, and UV coordinates. Given a text prompt, each view is painted using a depth-aware text-to-image diffusion model \cite{StableDiffusionModel}, guided by a pre-trained ControlNet \cite{ControlNet} that provides depth information\footnote{\href{https://huggingface.co/ashawkey/control_v11e_sd15_depth_aware_inpaint/tree/main}{https://huggingface.co/ashawkey/control\_v11e\_sd15\_depth\_aware\_inpaint}}. This ensures the generated textures align with both the text and depth (geometry) information. We repeat this process iteratively across all viewpoints until the full 3D surface is painted. Due to the stochastic nature of diffusion models, this can result in inconsistent textures and visible seams. To address this, we adopt a dynamic partitioning strategy inspired by \cite{TEXTure, Text2Tex}, using the \textit{view direction cosine} concept (see \cref{RemarkTextureGeneration}) to segment each view into three regions: (i) \textit{generate} – newly visible and unpainted areas, (ii) \textit{refine} – previously painted areas now seen from a better angle (higher cosine), and (iii) \textit{keep} – already well-painted regions that require no update. This partitioning guides the rendering and repainting process to maintain visual consistency. However, this partitioning and certain other operations in the texture generation step are not inherently differentiable, so we introduce mathematical modifications to ensure differentiability. These modifications are essential not only for making the pipeline differentiable, enabling the training of the diffusion model, but also for incorporating gradient information from each component of the texture-generation process. This ultimately makes the entire pipeline aware of the geometry of the mesh being textured. For more on the differentiable mathematical formulation, we refer the reader to \cref{RemarkMathematicTexGen} and \cref{ModificationsTextureGeneration}.
    }

    \subsection{Texture Rewards Learning}
    \label{TextureEnahncementStep}
    {   
        The texture image, obtained from the \textit{texture generation} step, serves as an input to the next step, \textit{texture reward learning}, to be evaluated and then enhanced according to a differentiable reward. The goal is to fine-tune the pre-trained stable diffusion model's parameters $\theta$ such that the texture images maximize a differentiable reward function $r$. The maximization problem can then be formulated as: 
        
        \vspace{-15pt}
        \begin{equation}            J(\boldsymbol{\theta})=\mathbb{E}_{\mathbf{c} \sim p_{\mathbf{c}}}\left[r\left(\operatorname{TexGen}\left(\boldsymbol{\theta}, \mathbf{c}, \mathbf{v_{gen}}\right), \mathbf{c}\right)\right]
        \label{TextureOptimizationEquation}
        \end{equation}        
        \vspace{-15pt}

        \noindent where $r$ is a differentiable reward  function, $\operatorname{TexGen}(\boldsymbol{\theta}, \mathbf{c}, \mathbf{v_{gen}})$ represents the iterative texturing process from different viewpoints $v_{gen}$ (explained in \cref{TextureGenerationStep} and \cref{RemarkTextureGeneration}), and $c$ is the text prompt randomly sampled from the text dataset $p_c$. To solve \cref{TextureOptimizationEquation}, instead of maximizing the reward function, we consider the negative of it as a loss function $L(\theta) = - J(\theta)$. Then, we compute $-\nabla_{\boldsymbol{\theta}}r\left(\operatorname{TexGen}\left(\boldsymbol{\theta}, \mathbf{c}, \mathbf{v_{gen}}\right), \mathbf{c}\right)$ and update $\theta$ using gradient-based optimization algorithms and the following update rule: $\theta \leftarrow \theta + \eta\nabla_{\boldsymbol{\theta}}r\left(\operatorname{TexGen}\left(\boldsymbol{\theta}, \mathbf{c}, \mathbf{v_{gen}}\right), \mathbf{c}\right)$. However, naively optimizing the $\theta$ would results in requiring a massive amount of memory which makes the fine-tuning process impractical. This is because of the fact that computing the gradients $\nabla_{\boldsymbol{\theta}}r$ requires backpropagation through multiple diffusion steps and multiple camera viewpoints which stems from the texturing process $\operatorname{TexGen}(\boldsymbol{\theta}, \mathbf{c}, \mathbf{v_{gen}})$. Consequently, this necessitates storing all the intermediate variables during the multi-view texturing process. More specifically, each step in the texturing process involves at least 10 diffusion steps, and with 8–10 total texturing steps (corresponding to the number of camera viewpoints), this results in approximately 100 diffusion steps overall. Each step requires around 3 GB of GPU memory, leading to a total memory demand of roughly 300 GB to train the texture-generation pipeline. Such hardware requirements are nearly infeasible, especially in typical academic environments. To alleviate this issue, we leverage two main approaches to reduce the memory used by our method during the training stage: 1) \textit{low-rank adaptation (LoRA)} \cite{LoRA}, and 2) \textit{gradient checkpointing} \cite{GradientCheckpointing}. Applying LoRA to the U-Net in the latent diffusion model reduces trainable parameters by roughly a factor of 1000 compared to full fine-tuning. Additionally, two levels of activation checkpointing keep memory usage constant across views and diffusion steps, enabling higher-resolution renders and more diffusion steps. Further memory-saving details are provided in \cref{RemarksTexturePreferenceLearning}.

        \vspace{-5pt}
        
        \subsection{Geometry-Aware Reward Design}
        \label{GeometryAwareRewardDesign}
        {            
            \vspace{-2pt}

             Our differentiable framework enables the design and implementation of task-specific reward functions that encode human preferences, operating in either the 2D texture image domain or directly on the 3D mesh surface, to steer texture generation toward the desired outcome. To illustrate its versatility, we introduce five novel geometry-aware reward functions. First, we present two simple rewards, \textit{geometry-aware texture colorization} and \textit{texture features emphasis}, to verify that our pipeline reliably fulfills each reward’s intended objective. We then propose three rewards with potential practical applications, \textit{geometry–texture alignment}, \textit{symmetry-aware texture generation}, and \textit{camera-pose-aware texture generation}, to demonstrate the pipeline’s capacity for producing high-fidelity, realistic, and task-specific textures on 3D shapes. Detailed descriptions of each reward follow in the subsections below and \cref{AttemptsToOvercomeRewardHacking}. For the \textit{geometry-aware colorization} reward, we refer the reader to \cref{GeometryGuidedTextureColorizationReward}.
            
            \begin{algorithm}[h]
                \caption{Differentiable 3D Texture Reward Learning}
                \label{alg:fine-tuning}
                \resizebox{0.9\columnwidth}{!}{%
                \begin{minipage}{\columnwidth}
                    \begin{algorithmic}[1]
                        \REQUIRE Diffusion steps $T$, training iterations $E$, prompt dataset $p_c$, viewpoints for generation $V_{gen}$ and evaluation $V_{eval}$, pre-trained model $\epsilon_\theta$ and diffusion model weights $\theta$
                        \FOR{$e = 1$ \textbf{to} $E$}
                            \begin{genblock}
                            \STATE \texttt{\# Step 1: Texture Generation}
                            \STATE Sample a prompt $c \sim p_c$, and sample $x_T \sim \mathcal{N}(\mathbf{0}, \mathbf{I})$.
                            \STATE Initialize 3D asset parameters $a$: texture $albedo$, weight $cnt$, view cache $viewcos$
                            \FOR{$v_{gen}$ \textbf{in} $V_{gen}$}
                                \STATE $h^i = \mathrm{DiffRender}(a, v_{gen})$
                                \FOR{$t = T$ \textbf{to} $1$}
                                    \STATE $y^i_{t-1} = \mathrm{ControlNet}(h^i, t, c)$
                                    \STATE $x^i_{t-1} = \mu(x^i_{t}, y^i_{t-1}, t, c) + \sigma_t z,\;\; z \sim \mathcal{N}(\mathbf{0}, \mathbf{I})$
                                \ENDFOR
                                \STATE $\mathrm{Inpaint3DAsset}(a, v_{gen})$
                                \STATE $albedo \leftarrow \mathrm{Update3DAsset}(a, v_{gen})$
                            \ENDFOR
                            \end{genblock}
            
                            \begin{learnblock}
                            \STATE \texttt{\# Step 2: Reward Learning}
                            \IF{3D Reward}
                                \FOR{$v_{eval}$ \textbf{in} $V_{eval}$}
                                    \STATE $h^i = \mathrm{DiffRender}(a, v_{eval})$
                                    \STATE $reward = reward + r(h^i)$
                                \ENDFOR
                                \STATE $reward = reward / len(V_{eval})$
                                \STATE $g = -\nabla_{\theta} reward$
                            \ELSIF{Texture Reward}
                                \STATE $g = -\nabla_{\theta} r(albedo)$
                            \ENDIF
                            \STATE $\theta \leftarrow \theta - \eta g$
                            \end{learnblock}
                        \ENDFOR
                        \STATE \textbf{return} $\theta$
                    \end{algorithmic}
                \end{minipage}
                }
            \end{algorithm}

            \noindent \textbf{Texture Features Emphasis Reward.}\label{TextureFeaturesEmphasisReward}
            {
                This reward encourages texture patterns that emphasize 3D surface structure while maintaining perceptual richness through color variation. Specifically, it strengthens texture variations in high-curvature regions to enhance capturing geometric structures. Let $T(x,y) \in [0,1]^3$ denote the RGB texture at pixel $(x,y)$ and $C(x,y) \in [0,1]$ be the normalized curvature map in UV space. The \textit{texture-curvature alignment} term is mathematically defined as the negative mean squared error (MSE):
                
                \vskip -0.18in
                \begin{equation}
                    R_{magnitude}=-\frac{1}{N} \sum_{x, y}({\nabla} {\mathbf{T}}(x, y)-{C}(x, y))^2
                \end{equation}
                \vskip -0.1in
                
                \noindent where $\nabla \mathrm{T}(x, y)=\sqrt{\left\|\frac{\partial \mathrm{T}}{\partial x}(x, y)\right\|^2+\left\|\frac{\partial \mathrm{T}}{\partial y}(x, y)\right\|^2}$ is the texture gradient magnitude and $N$ is the total number of pixels in the texture image. Further details about this reward are provided in \cref{RemarksTextureFeaturesEmphasisReward}.
            }

             \noindent \textbf{Geometry-Texture Alignment Reward.}
            \label{GeometryTextureAlignmentReward}
            {
                To encourage texture image features (e.g. edges) to be aligned with the geometry of a 3D object, we introduce a differentiable reward function that aligns texture gradient vectors with principal curvature directions on the mesh surface (see \cref{RemarkPrincipalCurvature} for details). These curvature vectors encode the surface’s bending behavior, and aligning texture gradients with them results in textures that are both visually and semantically consistent with the underlyinggeomtery of the 3D input mesh. More specifically, for each  point in the UV space (texture) obtained by projecting 3D mesh vertices into 2D texture space, we compute the alignment as the squared dot product (cosine similarity) between the normalized texture gradient and corresponding normalized minimum curvature direction vector:
                
                \vspace{-22pt}
                \begin{equation}
                    \label{cosine_similarity}
                    R_1=\frac{1}{N} \sum_{i=1}^N(\overrightarrow{\mathbf{g}}_i \cdot \overrightarrow{\mathbf{c}}_i)^2
                \end{equation}
                \vspace{-8pt}

                \noindent where $N$ is the number of UV coordinates, $\overrightarrow{\mathbf{g}}_i$ and $\overrightarrow{\mathbf{c}}_i$ are the texture gradient and minimum principal curvature at each UV coordinate, respectively. This alignment works because curvature directions capture the intrinsic anisotropy of the surface, guiding texture flow along perceptually meaningful geometric features such as ridges, valleys, and folds. Note that minimum and maximum curvatures are a vertex-level attribute defined per vertex on the 3D mesh surface, whereas texture gradients are defined per pixel in the 2D UV space. Therefore, to guide texture generation using surface-aware vector fields and to compute cosine similarity between curvature vectors and texture gradients (\cref{cosine_similarity}), both must be in the same space. To address this, we project the minimum principal curvature direction vectors, defined at each vertex on the 3D mesh, into the corresponding 2D UV space. Unlike scalar quantities like mean curvature, vector fields require special process to preserve both direction and tangency. This process involves the following steps. 
                
                \noindent \textbf{(\RNum{1}) Computing principal curvature directions.} We use the method from libigl \cite{libigl} to compute the minimum principal curvature direction $\mathbf{d}_i \in \mathbb{R}^3$ at each vertex $\mathbf{v}_i \in \mathbb{R}^3$ of the mesh. Each direction vector lies in the tangent plane of the surface at its vertex. 
                
                \noindent \textbf{(\RNum{2}) Projecting to the tangent plane.}
                In the next step, the calculation is restricted to the tangent plane of $\mathbf{v}_i$ so that we project one-ring adjacent vertices of $\mathbf{v}_i$ to the tangent plane. Although the direction vectors are already tangent to the surface by construction, we explicitly project them onto the local tangent plane to remove residual normal components (due to numerical noise). For each vertex: $\mathbf{d}_i^{\tan }=\mathbf{d}_i-\left(\mathbf{d}_i \cdot \mathbf{n}_i\right) \mathbf{n}_i$ where $\mathbf{n}_i$ is the surface normal at vertex $i$. 
                
                \noindent \textbf{(\RNum{3}) Tracing vector to find triangle containment.} For each projected vector $\mathbf{d}_i^{\tan}$, we trace it forward from the vertex $v_i$ to obtain a new point: $\mathbf{p}_i=\mathbf{v}_i+\frac{1}{\lambda_i} \cdot \mathbf{d}_i^{\tan }$ where $\lambda_i$ is a normalization constant to ensure the traced vector remains within a local triangle. We then search among the projected triangles incident to $v_i$ to find one in which the point $\mathbf{p}_i$ lies. Once found, we compute the barycentric coordinates $\beta_i = [\beta_1, \beta_2, \beta_3]$ of $\mathbf{p}_i$ relative to the triangle's projected vertices. 
                
                \noindent \textbf{(\RNum{4}) Converting to UV coordinates.} Using the barycentric weights and the corresponding UV triangle $(\mathbf{u}_1, \mathbf{u}_2, \mathbf{u}_3)$, we map the end of the vector to UV space: $\mathbf{u}_{\text{end}} = \beta_1\mathbf{u}_1 + \beta_2\mathbf{u}_2 + \beta_3\mathbf{u}_3$. The UV origin $\mathbf{u}_{\text{orig}}$ is determined from the texture coordinate associated with vertex $\mathbf{v}_i$. The 2D texture-space direction vector is then: 

                \vspace{-10pt}
                \begin{equation}
                    \mathbf{d}_i^{\mathbf{u v}}=\frac{\mathbf{u}_{\text {end }}-\mathbf{u}_{\text {orig }}}{\left\|\mathbf{u}_{\text {end }}-\mathbf{u}_{\text {orig }}\right\|}
                \end{equation}
                \vspace{-5pt}
        
                \noindent \textbf{(\RNum{5}) Populating texture coordinates.} Since a single vertex may be associated with multiple UV coordinates (due to UV seams), the final $ \mathbf{d}_i^{\mathbf{u v}}$ is assigned to all associated texel indices. This step builds a complete UV-space vector field aligned with curvature directions. 
                
                \noindent \textbf{(\RNum{6}) Post-processing and smoothing.} Some vertices may not yield a valid projection due to degenerate geometry, occlusion, or sharp creases. For these cases, we apply Laplacian smoothing (iterative neighbor averaging) technique using a texture-space adjacency graph. After resolving missing values, we apply several iterations of vector field smoothing to reduce visual noise while preserving directionality: $\left.\mathbf{d}_i^{\text {uv }} \leftarrow \text { mean(neighbors of } i\right)$. This vector field in UV space is then used to supervise or visualize alignment of learned texture patterns with surface geometry using \cref{cosine_similarity}.                
            }

             \noindent \textbf{Symmetry-Aware Texture Generation Reward.}
            \label{SymmetryAwareTextureGenerationReward}
            {
                We propose a differentiable reward function that encourages texture patterns to be symmetric across mirrored UV coordinates while maintaining visual richness through color variation. We first identify the mesh’s intrinsic plane of symmetry using Principal Component Analysis (PCA) \cite{PCA1, PCA2}, then compute mirror pairs of surface points across this plane, project these pairs into the 2D UV domain for use in a symmetry-aware reward function, and finally compute the symmetry reward. The four steps are outlined in the following.
                
                \noindent \textbf{(\RNum{1}) Symmetry plane estimation via PCA.} 
                {
                    To estimate the mesh's dominant symmetry plane, we apply principal component analysis (PCA) to the 3D vertext coordinates. Let $\{{v_i} \in\mathbb{R}^3\}_{i=1}^N $ denote the 3D vertext positions of the mesh. We compute the centroid of the mesh as $\mathbf{c}=\frac{1}{N} \sum_{i=1}^N \mathbf{v}_i$. Next, we construct the covariance matrix of the centered vertices:        
                    $\Sigma=\frac{1}{N} \sum_{i=1}^N\left(\mathbf{v}_i-\mathbf{c}\right)\left(\mathbf{v}_i-\mathbf{c}\right)^{\top}$. We then perform eigen-decomposition of $\Sigma$, yielding eigenvalues $\lambda_1 \leq \lambda_2 \leq \lambda_3 \leq$ and corresponding orthonormal eigenvectors $e_1, e_2, e_3$. The eigenvector $n=e_1$ associated with the smallest eigenvalue is used to define the normal vector of the symmetry plane, as it corresponds to the direction of least variance in the point cloud (vertices). The symmetry plane $\Pi$ is then defined as the following set of points:
                    \begin{equation}
                        \Pi=\left\{\mathrm{x} \in \mathbb{R}^3 \mid(\mathrm{x}-\mathrm{c}) \cdot \mathrm{n}=0\right\}
                    \end{equation}
                }
                \noindent \textbf{(\RNum{2}) Reflecting vertices across the symmetry plane.}
                {
                    To identify symmetric correspondences, we first compute the signed distance of each vertex to the plane: $d_i=\left(\mathbf{v}_i-\mathbf{c}\right) \cdot \mathbf{n}$. We keep only the vertices where $d_i > 0$, i.e., those lying on one side of the plane, to avoid duplication. Each selected $v_i$ is then reflected across the plane to compute its mirrored position $\mathbf{v}_i^{\text {mirror }} \in \mathbb{R}^3$ using: $\mathbf{v}_i^{\text {mirror }}=\mathbf{v}_i-2\left[\left(\mathbf{v}_i-\mathbf{c}\right) \cdot \mathbf{n}\right] \mathbf{n}$.
                }
            
                \noindent \textbf{(\RNum{3}) Projecting 3D mirror points into UV space.}
                {
                    To obtain the UV coordinates corresponding to $\mathbf{v}_i^{\text {mirror }}$, we perform a closest-point query on the mesh surface to find the nearest triangle containing the surface projection of $\mathbf{v}_i^{\text {mirror }}$. Given the triangle's 3D vertices $\mathbf{v}_1, \mathbf{v}_2, \mathbf{v}_3$, and corresponding UV coordinates $\mathbf{u}_1, \mathbf{u}_2, \mathbf{u}_3 \in [0,1]^2$, we compute barycentric coordinates $(\beta_1, \beta_2, \beta_3)$ of the projected point and obtain the UV coordinates of the mirror as: $\mathbf{u}_i^{\text {mirror }}=\beta_1 \mathbf{u}_1+\beta_2 \mathbf{u}_2+\beta_3 \mathbf{u}_3$. We store the UV coordinate $\mathbf{u}_i$ of the original vertex along with $\mathbf{u}_i^{\text {mirror}}$ and form a pair $(\mathbf{u}_i, \mathbf{u}_i^{\text {mirror}})$ used in our symmetry-awared reward.
                }
            
                \noindent \textbf{(\RNum{4}) Symmetry-aware texture reward design.}
                {
                    Given a set of UV coordinate pairs \( \{ (\mathbf{u}_i, \mathbf{u}_i^{\text{mirror}}) \}_{i=1}^M \), obtained by projecting symmetric 3D vertex pairs into the UV domain, we define a differentiable reward function that encourages symmetry in the generated texture. Let \( \mathcal{T} \in \mathbb{R}^{3 \times H \times W} \) denote the RGB texture defined in UV space. For each UV pair, we sample RGB values via bilinear interpolation:
                    \begin{equation}
                        \mathcal{T}_i = \text{Sample}(\mathcal{T}, \mathbf{u}_i), \quad \mathcal{T}_i^{\text{mirror}} = \text{Sample}(\mathcal{T}, \mathbf{u}_i^{\text{mirror}})
                    \end{equation}
                    
                    \noindent where \( \mathcal{T}_i, \mathcal{T}_i^{\text{mirror}} \in \mathbb{R}^3 \) represent the RGB values at each location. Sampling is implemented using the differentiable \texttt{grid\_sample} operation in Pytorch \cite{Pytorch} to ensure gradient flow during training. To encourage symmetry, we minimize the MSE between original and mirrored samples:

                    \vspace{-12pt}
                    \begin{equation}
                        \mathcal{L}_{\text{symmetry}} = \frac{1}{M} \sum_{i=1}^M \left\| \mathcal{T}_i - \mathcal{T}_i^{\text{mirror}} \right\|_2^2
                    \end{equation}
                    \vspace{-12pt}
                    
                    We define the symmetry reward as the negative of this loss:

                    \vspace{-12pt}
                    \begin{equation}
                        \mathcal{R}_{\text{symmetry}} = - \mathcal{L}_{\text{symmetry}}
                    \end{equation}
                    \vspace{-12pt}
                    
                    \noindent To discourage trivial (e.g., grayscale) solutions, similar to \textit{texture features emphasis} reward (\cref{TextureFeaturesEmphasisReward}), we augment the symmetry reward with a colorfulness term using \cref{Colorfulness_reward}. The total reward is a weighted combination of the symmetry alignment and colorfulness terms:

                    \vspace{-10pt}
                    \begin{equation}
                        \mathcal{R}_{4} = \alpha_{\text{sym}} \cdot \mathcal{R}_{\text{symmetry}} + \alpha_{\text{color}} \cdot \mathcal{R}_{\text{color}}
                    \end{equation}
                    \vspace{-10pt}
                    
                    \noindent We use \( \alpha_{\text{sym}} = 1.0 \) and \( \alpha_{\text{color}} = 0.05 \) in our experiments.
                }
            }

            \noindent \textbf{Camera-Pose-Aware Texture Generation Reward.}
            {
                Iteratively painting the 3D object from different camera viewpoints (see \cref{TextureGenerationStep}) requires careful tuning of camera positions and orientations in 3D space, so that the entire mesh surface is covered during texturing. This manual tuning is both arduous and time‑consuming, and must be repeated for every object. In fact, most texturing algorithms, such as TEXTure \cite{TEXTure} and InTeX \cite{InTex} suffer from this limitation. To address this shortcoming, we propose to learn the camera viewpoints during training by treating the camera’s azimuth and elevation angles as trainable parameters. Specifically, at each iteration of the texturing process, we initialize the azimuth and elevation randomly, execute our texture‐generation step to render the object, and then evaluate each rendered view using the LAION aesthetic predictor \cite{LAION_Aesthetic_Score}. We average these aesthetic scores and back‑propagate the negative of this reward through our end‑to‑end differentiable pipeline. In this way, the framework learns to adjust camera parameters so as to maximize the average aesthetic reward. Intuitively, if a camera is poorly positioned, the resulting render is suboptimal and yields imperfect paintings from the depth‑aware diffusion model. By contrast, optimizing the viewpoints steers the model toward those positions and orientations that best capture a given object. Formally, we reformulate \cref{TextureOptimizationEquation} as
                
                \vspace{-15pt}
                \begin{equation}
                J(\boldsymbol{\phi}) = \mathbb{E}_{\mathbf{c} \sim p{\mathbf{c}}}\bigl[r\bigl(\operatorname{TexGen}(\boldsymbol{\theta},\mathbf{c},\mathbf{v}_{\mathrm{gen}}(\boldsymbol{\phi})),\mathbf{c}\bigr)\bigr]
                \label{CameraPoseOptimizationEquation}
                \end{equation}
                \vspace{-15pt}
                
                \noindent where $r$ denotes the LAION aesthetic reward \cite{LAION_Aesthetic_Score}, and $\operatorname{TexGen}(\boldsymbol{\theta},\mathbf{c},\mathbf{v}{\mathrm{gen}}(\boldsymbol{\phi}))$ corresponds to the texture‐generation step of \cref{TextureGenerationStep}, with the diffusion model parameters $\boldsymbol{\theta}$ held fixed and the camera viewpoints $\mathbf{v}_{\mathrm{gen}}(\boldsymbol{\phi})$ learned instead of predefined. To solve \cref{CameraPoseOptimizationEquation}, instead of maximizing the aesthetic reward, we consider the negative of it as a loss function $L(\phi) = - J(\phi)$. Then, we compute $-\nabla_{\boldsymbol{\phi}}r\left(\operatorname{TexGen}\left(\boldsymbol{\theta}, \mathbf{c}, \mathbf{v_{gen}(\phi)}\right), \mathbf{c}\right)$ and update $\phi$ using gradient-based optimization algorithms and the following update rule: $\phi \leftarrow \phi + \eta\nabla_{\boldsymbol{\phi}}r\left(\operatorname{TexGen}\left(\boldsymbol{\theta}, \mathbf{c}, \mathbf{v_{gen}(\phi)}\right), \mathbf{c}\right)$. 

            }
        }
    }

    \begin{figure}[ht]
        \vspace{-5pt}
        \begin{center}
        \centerline{\includegraphics[scale=0.24]{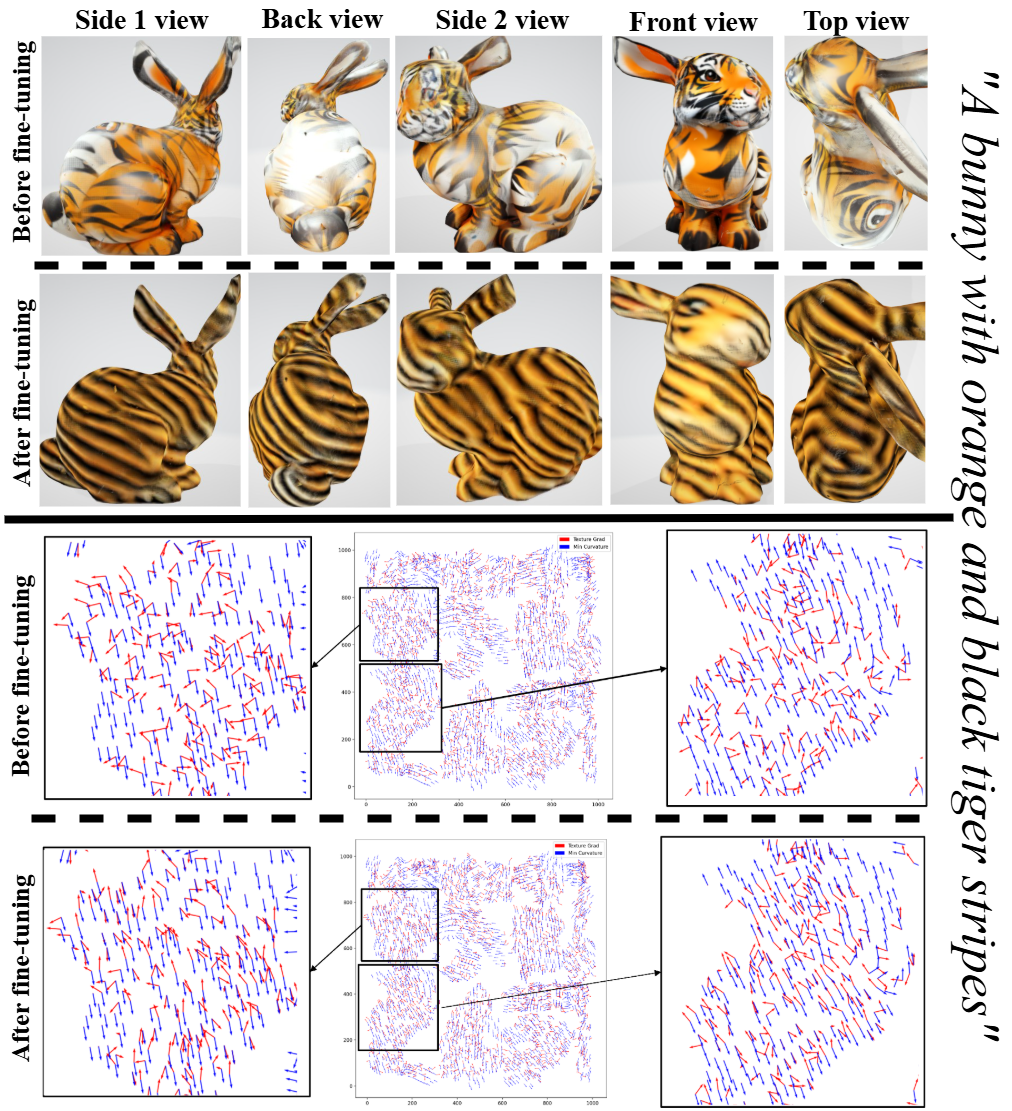}}
        \caption{Qualitative results of the geometry–texture alignment experiment on a rabbit (bunny) mesh. For both the pre‑fine‑tuning and post‑fine‑tuning cases, we render the bunny from several single viewpoints. In the bottom two rows, we visualize 2D‑projected minimum curvature vectors (blue) alongside texture gradient vectors (red) to highlight alignment. After fine‑tuning with our alignment reward, the texture patterns conform much more closely to the mesh’s curvature directions.
        \vspace{-35pt}}
        \label{CurvTexAlignQualitativeResultShort}
        \end{center}                
    \end{figure}
}
\section{Experiments and Results}
\label{sec:experiments_results}
{
    \vspace{-5pt}

    \begin{table}[t]
    \vspace{-10pt}
    \caption{\textbf{Top table:} A quantitative comparison between our proposed method (after fine-tuning) and InTeX \cite{InTex} (before fine-tuning) across five experiments (the top table). We repeat each experiment five times and report the mean and standard deviation values of rewards. GEO-TEX-ALIGN, GEO-TEX-COLOR, TEX-EMPHASIS, SYM-AWARE, and CAM-POSE-AWARE stand for geometry-texture alignment reward, geometry-guided texture colorization reward (\cref{GeometryGuidedTextureColorizationReward}), texture features emphasis reward, symmetry-aware, and camera-pose-aware texture generation reward, respectively. \textbf{Bottom table:} A user‐study with 40 participants performing 11 pairwise comparisons; one‐sided binomial tests show that our post‑learning textures are significantly preferred over InTeX outputs. \vspace{-15pt}}
    \label{QuantitativeResults_UserStudy}
    \vskip 0.15in
    \begin{center}
    \begin{small}
    \begin{sc}
    \scalebox{0.7}{\begin{tabular}{lcccr}
    \toprule
    Reward & Ours & InTeX \cite{InTex} \\
    \midrule
    Geo-Tex-Align    & \textbf{4.7984} $\pm$ 0.0791 & 3.7158 $\pm$ 0.0168 \\
    Geo-Tex-Color    & \textbf{0.6006} $\pm$ 0.0003 & -0.0256 $\pm$ 0.0195 \\
    Tex-Emphasis     & \textbf{-0.1106} $\pm$ 0.0002 & -0.1423 $\pm$ 0.0007 \\
    Sym-Aware        & \textbf{0.0114} $\pm$ 0.0093 & -0.0635 $\pm$ 0.0039 \\
    Cam-Pose-Aware        & \textbf{5.4035} $\pm$ 0.0747 & 4.0900 $\pm$ 0.2177 \\    
    \bottomrule
    User Preference Study & Preferred Fine-tuned &  Binomial Test (p-value) \\
    \midrule
    Geo-Tex-Align    & \textbf{98\%} & $1.8140 \times 10^{-29}$ \\
    Tex-Emphasis     & \textbf{93.9\%} & $2.3104 \times 10^{-16}$ \\
    Sym-Aware        & \textbf{88.86\%}  & $6.9346 \times 10^{-18}$ \\
    Cam-Pose-Aware   & \textbf{98\%} &  $1.8140 \times 10^{-28}$  \\
    \bottomrule
    \end{tabular}}
    \end{sc}
    \end{small}
    \end{center}
    \vskip -0.3in
    \end{table}
    
    To show the effectiveness of our approach, we perform four training experiments using the four proposed geomatry-aware reward functions in \cref{GeometryAwareRewardDesign}, each detailed in the following. We qualitatively and quantitatively compare our results against InTeX \cite{InTex}, TEXTure \cite{TEXTure}, Text2Tex \cite{Text2Tex}, and Paint3D \cite{Paint3D}, four well-established state-of-the-art methods in the field of texture generation. We report the qualitative results for each experiment in \cref{CurvTexAlignQualitativeResultShort}, and \cref{ThreeRewards_Results}, and quantitative results in \cref{QuantitativeResults_UserStudy}, \cref{QuantitativeResults_metrics}, and \cref{QuantitativeResults} (more results in \cref{AdditionalExperimentsResults}). 

    \noindent \textbf{Quantitative Comparison.}
    {
        We report a variety of widely-used automatic metrics to evaluate visual quality, image-text alignment and human preference proxies in \cref{QuantitativeResults_metrics}. Moreover, as shown at the top of \cref{QuantitativeResults_UserStudy}, we report the reward values for both our post-fine-tuned method and the pre-fine-tuned baseline (InTeX \cite{InTex}) across five experiments. Each experiment is repeated five times, and we report the mean and standard deviation of the resulting reward values. For each scene we render the textured mesh from multiple viewpoints. In \cref{QuantitativeResults_metrics}, for each method and scene we compute the selected scalar metric per rendered view, then average those per-view values to obtain a \textit{3D-scene-level} score for that method. Finally, we report the mean of those scene-level scores across scenes. As presented in \cref{QuantitativeResults_metrics}, our method consistently outperforms InTeX \cite{InTex}, TEXTure \cite{TEXTure}, Text2Tex \cite{Text2Tex}, and Paint3D \cite{Paint3D} numerically. Below we briefly define each metric and give the intended interpretation (higher = better or lower = better). \textbf{(i) Aesthetic (LAION predictor)} \cite{LAION_Aesthetic_Score}: an image-level aesthetic score produced by a learned aesthetic predictor trained using LAION-style data.  It is intended to correlate with general photographic/artistic quality and larger values indicate more aesthetically pleasing images. \textbf{(ii) ImageReward} \cite{ImageReward}: a learned reward model trained on human preferences that predicts which images humans prefer given a prompt or context. We use the ImageReward model as a proxy for human preference; larger scores indicate texture images more likely to be preferred by annotators. \textbf{(iii) HPSv2} (Human Preference Score v2) \cite{HPSv2}: an improved preference-model variant tuned to better correlate with human judgments on text-to-image outputs. Like other preference scorers, higher is better. \textbf{(iv) PickScore} \cite{PickScore}: a preference model trained on the \textit{Pick-a-Pic} dataset \cite{Pick-a-pic_dataset} to predict which of two candidate images a human would choose. We report the model’s scalar output aggregated across views. Higher is better. \textbf{(v) CLIPScore} \cite{CLIPScore}: the cosine similarity between image and text embeddings from CLIP \cite{CLIP}. We compute per-view CLIP similarity between the render image and the prompt and average across views. This measures image–text alignment and higher values indicate better alignment. \textbf{(vi) Inference time}: average time to produce a final textured mesh (or to run the texture generation/inference procedure for one object) reported in seconds. Lower is better; all timing measurements were performed on a machine equipped with a single NVIDIA RTX 4090 GPU. In the following, we detail each experiment and provide a qualitative comparison against three strong baselines (see \cref{ThreeRewards_Results} in \cref{AdditionalExperimentsResults}): InTeX \cite{InTex}, TEXTure \cite{TEXTure}, Text2Tex \cite{Text2Tex}, and Paint3D \cite{Paint3D}.
            
        \vspace{-8pt}

        \begin{table*}[t]
            \centering
            \vspace{-12pt}
            \caption{Quantitative comparison of our method against baselines across multiple automatic metrics. We report higher-is-better ($\uparrow$) scores for Aesthetic \cite{LAION_Aesthetic_Score}, ImageReward \cite{ImageReward}, HPSv2 \cite{HPSv2}, PickScore \cite{PickScore}, and CLIPScore \cite{CLIPScore}, and a lower-is-better ($\downarrow$) measure for inference time. Across these metrics, our method consistently outperforms all four baselines.}
            \vspace{-5pt}
            \resizebox{\textwidth}{!}{
            \begin{tabular}{lcccccccc}
            \toprule
            \textbf{Method} & 
            \textbf{Aesthetic} $\uparrow$ & 
            \textbf{ImageReward} $\uparrow$ & 
            \textbf{HPSv2} $\uparrow$ & 
            \textbf{PickScore} $\uparrow$ & 
            \textbf{CLIPScore} $\uparrow$ & 
            \textbf{Inference Time (sec)} $\downarrow$\\
            \midrule
            TEXTure \cite{TEXTure} & 4.3910 $\pm$ 0.0328 & -1.2245 $\pm$ 0.1199 & 0.1728 $\pm$ 0.0031 & 20.0719 $\pm$ 0.1083 & 0.3142 $\pm$ 0.0051 & 150 \\
            Text2Tex \cite{Text2Tex} & 4.4454 $\pm$ 0.0387 & -1.5088 $\pm$ 0.0823 & 0.1769 $\pm$ 0.0052 & 19.6886 $\pm$ 0.1862 & 0.2900 $\pm$ 0.0054 & 450  \\
            InTeX \cite{InTex} & 4.7467 $\pm$ 0.0381 & -1.0859 $\pm$ 0.1469 & 0.1879 $\pm$ 0.0033 & 19.8832 $\pm$ 0.2368 & 0.3077 $\pm$ 0.0055 & 15  \\
            Paint3D \cite{Paint3D} & 4.7192 $\pm$ 0.0388 & -1.6621 $\pm$ 0.0977 & 0.1549 $\pm$ 0.0089 & 18.8966 $\pm$ 0.2798 & 0.2798 $\pm$ 0.0065 & 30  \\
            \midrule
            Ours - Cam-Pose-Aware Reward & 4.9328 $\pm$ 0.0271 & -0.0479 $\pm$ 0.1748 & 0.2095 $\pm$ 0.0058 & 19.7609 $\pm$ 0.1573 & 0.3003 $\pm$ 0.0057 & 15 \\
            Ours - Geo-Tex-Align Reward & 4.7722 $\pm$ 0.0324 & -0.1758 $\pm$ 0.1136 & 0.2479 $\pm$ 0.0034 & \textbf{21.5022} $\pm$ 0.1293 & \textbf{0.3367} $\pm$ 0.0041 & 15 \\
            Ours - Sym-Aware Reward & 4.9473 $\pm$ 0.0309 & \textbf{0.0063} $\pm$ 0.0901 & 0.2118 $\pm$ 0.0031 & 20.8361 $\pm$ 0.0444 & 0.3076 $\pm$ 0.0032 & 15 \\
            Ours - Tex-Emphasis Reward & \textbf{5.0308} $\pm$ 0.0376 & -0.5960 $\pm$ 0.1694 & \textbf{0.2534} $\pm$ 0.0027 & 20.8822 $\pm$ 0.1125 & 0.3158 $\pm$ 0.0028 & \textbf{15} \\
            \bottomrule
            \end{tabular}
            }
            \vspace{-10pt}
            \label{QuantitativeResults_metrics}            
        \end{table*}
    }

    \vspace{5pt}
    \noindent \textbf{Geometry-Texture Alignment.}
    {        
        The objective of this task is to generate texture images whose patterns are aligned with the surface geometry of a 3D mesh object, represented by its minimal curvature directions. \cref{CurvTexAlignQualitativeResultShort} presents qualitative results on a rabbit mesh textured using the textual prompt "A bunny with orange and black tiger stripes". For both the pre-fine-tuning and post-fine-tuning cases, we render the bunny from several single viewpoints. In the bottom two rows, we visualize 2D-projected minimum curvature vectors (blue) alongside texture gradient vectors (red) to highlight alignment. As illustrated, our method successfully guides texture patterns to follow the underlying geometric cues, a capability that InTex \cite{InTex} lacks (more results in \cref{CurvTexAlign_Qualitative}). Compared to InTeX, which uses a pre-trained model without fine-tuning, our approach exhibits an interesting capability: the generation of repetitive texture patterns that align with the curvature structure. This behavior arises from the differentiable sampling step we introduce during the geometry-texture alignment stage, where texture gradients are sampled at the same spatial locations as curvature vectors. By consistently associating these positions with edge-like features in the texture, the model learns to reinforce structure-aware repetition. Additional analysis of this effect are provided in \cref{EmergenceRepetitivePatterns}.
    }


    \noindent \textbf{Texture Features Emphasis.}
    {
        This task aims to learn texture images with salient features (e.g., edges) emphasized at regions of high surface bending, represented by the magnitude of mean curvature. This promotes texture patterns that highlight 3D surface structure while preserving perceptual richness through color variation. \cref{ThreeRewards_Results} (right) shows qualitative results on a rabbit mesh with brick patterns. As illustrated, our method enhances texture features, such as edges and mortar, in proportion to local curvature, a capability all baselines lack, often resulting in pattern-less (white) areas, particularly on the back and head of the rabbit (more results in \cref{TextureFeatureEmphasis_Qualitative}).

    }

    \noindent \textbf{Symmetry-Aware Texture Generation.}
    {
        The goal of this experiment is to encourages texture consistency across symmetric regions of a 3D object (see \cref{SymmetryAwareTextureGenerationReward}). We evaluate this by training our pipeline using the symmetry reward on a balloon mesh object. \cref{ThreeRewards_Results} (middle)  presents qualitative results for the experiments with symmetry-aware reward. For each viewpoint, we present the rendered mesh with a vertical dashed line indicating the symmetry axis. Compared to the baselines, our method produces textures that exhibit consistent patterns across symmetric regions of the mesh, confirming the reward’s ability to enforce symmetry. In contrast, without symmetry supervision, the textures on opposite sides often diverge significantly (more results in \cref{Symmetry_Qualitative}).

        \noindent \textbf{Camera-Pose-Aware Texture Generation.}
        {
            The objective of this experiment is to learn optimal camera viewpoints such that, when the object is rendered and textured from these views, the resulting texture maximizes the average aesthetic reward. Consequently, by maximizing this reward, the model becomes invariant to the initial camera positions: regardless of where the cameras start, the training will adjust their azimuth and elevation to surround the 3D object in a way that yields high‑quality, aesthetically pleasing textures. \cref{ThreeRewards_Results} (left) presents qualitative results on three different meshes, a female head, a miniature dragon, and a male head, comparing our fine‑tuning approach against baselines. All methods share identical initial camera parameters; once initialized, each algorithm proceeds to paint the object. As shown in \cref{ThreeRewards_Results}, our method consistently outperforms all baselines, producing high‑fidelity, geometry‑aware textures across all shapes. This robustness stems from the aesthetic‑reward signal guiding camera adjustment during texture reward learning. In contrast, InTeX \cite{InTex}, TEXTure \cite{TEXTure}, and Paint3D \cite{Paint3D} remain sensitive to poorly chosen initial viewpoints and thus often fail to generate coherent, high‑quality textures under suboptimal camera configurations. More results on this experiment are provided in \cref{CameraPoseAwareTexGen_Qualitative}.            
        }

        \noindent \textbf{User Study.}
        {
            We conducted a user‐preference study with 40 participants, each of whom completed 11 pairwise comparisons between textured 3D shapes generated by our method and by InTeX \cite{InTex}. Participants rated visual quality according to four criteria: fidelity, symmetry, overall appearance, and text‑and‑texture pattern alignment. Specifically, for the three reward variants, cam‐pose‐aware texture generation, symmetry‐aware texture generation, and geometry–texture alignment, we asked three questions apiece, and for the texture‐features‐emphasis reward we asked two questions. We then applied one‐sided binomial tests to each experiment to obtain p‑values (see bottom of \cref{QuantitativeResults_UserStudy}). As the table shows, the post‑reward‐learning results are significantly preferred over the pre‑learning outputs. Examples of questions in our user‐study questionnaire are shown in \cref{UserStudyFigure}.
            \vspace{-15pt}
        }
    }
}

\section{Conclusions}
\label{sec:conclusion}
{

    \vspace{-5pt}

    This study presents an end-to-end differentiable reward learning framework for 3D texture generation, enabling human preferences, expressed as differentiable reward functions, to be back-propagated through the entire 3D generative pipeline. This makes the process inherently geometry-aware. To demonstrate its effectiveness, we introduce three geometry-aware reward functions that guide texture generation to align closely with the input mesh geometry. The framework is broadly applicable, supporting any differentiable feedback defined in either 2D texture space or directly on the 3D surface, and is extensible to tasks beyond texture generation. As future work, we envision applying this approach to general 3D content creation, including joint optimization of geometry and texture.

}

\section*{Acknowledgements}
{
    We acknowledge funding from the Mila Tech Transfer Grant with Silicolabs and internal funds from Concordia University. Computational resources were provided by Calcul Québec and the Digital Research Alliance of Canada.
}

{
    \small
    \bibliographystyle{ieeenat_fullname}
    \bibliography{main}
}

\clearpage
\appendix
\setcounter{page}{1}
\maketitlesupplementary

\section{Remarks on The Texture Generation Step}
\label{RemarkTextureGeneration}
{
    \begin{figure}[h]
        \vskip -0.1in
        \begin{center}
        \centerline{\includegraphics[scale=0.27]{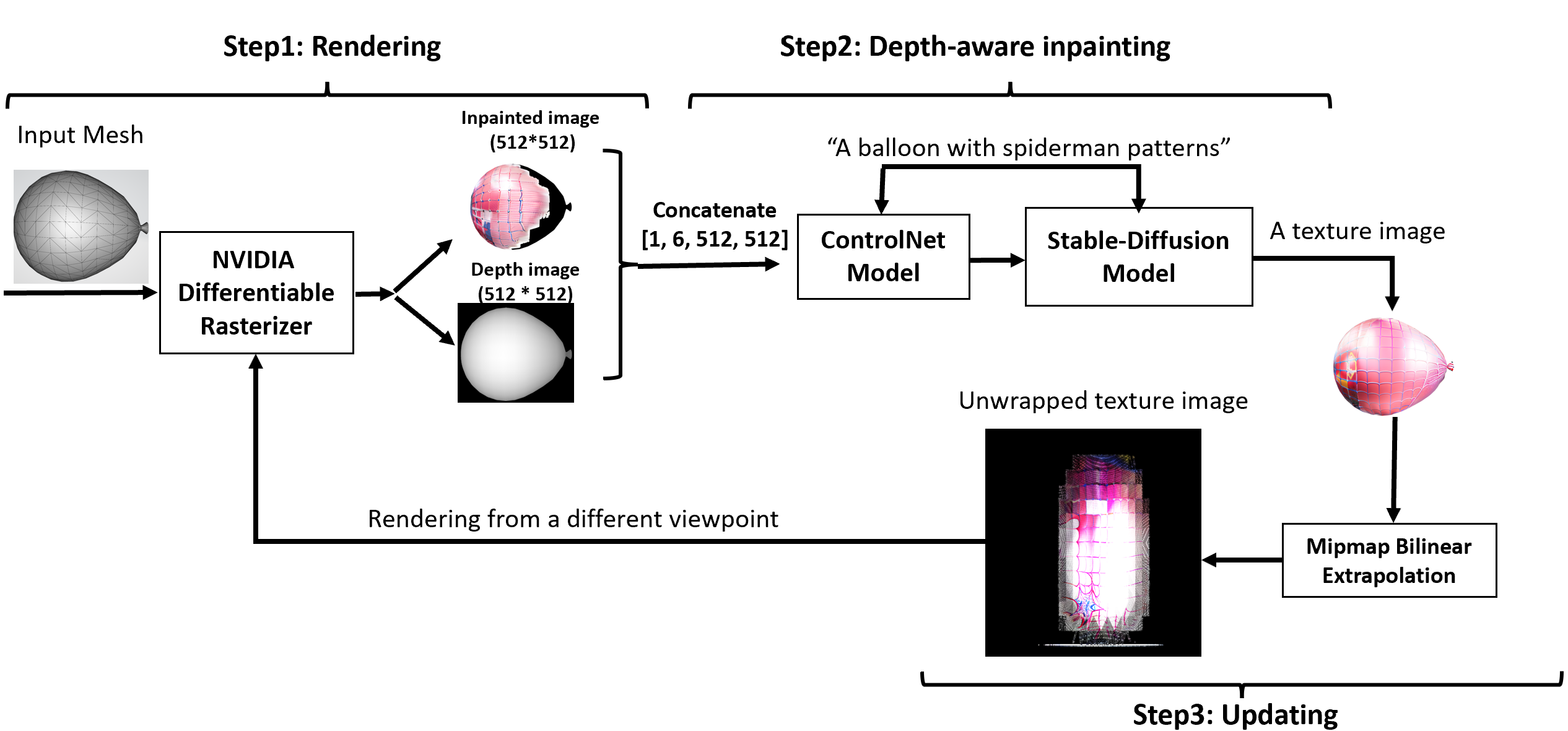}}
        \caption{Visualization of three main stages in the \textit{texture generation}: (i) rendering, render the object from a camera viewpoint using a differentiable renderer and extract a rendering buffer, (ii) depth-aware painting, given a text prompt, each viewpoint is painted using a depth-aware text-to-image diffusion model, guided by a pre-trained ControlNet that provides depth information. This ensures the generated textures align with both the text and depth (geometry) information, and (iii) update the final texture. We repeat this process iteratively across all camera viewpoints until the full 3D surface is painted.}
        \label{TextureGenerationSteps}
        \end{center}
        \vskip -0.1in
    \end{figure}


    \begin{figure}[h]            
            \begin{center}
            \centerline{\includegraphics[scale=0.22]{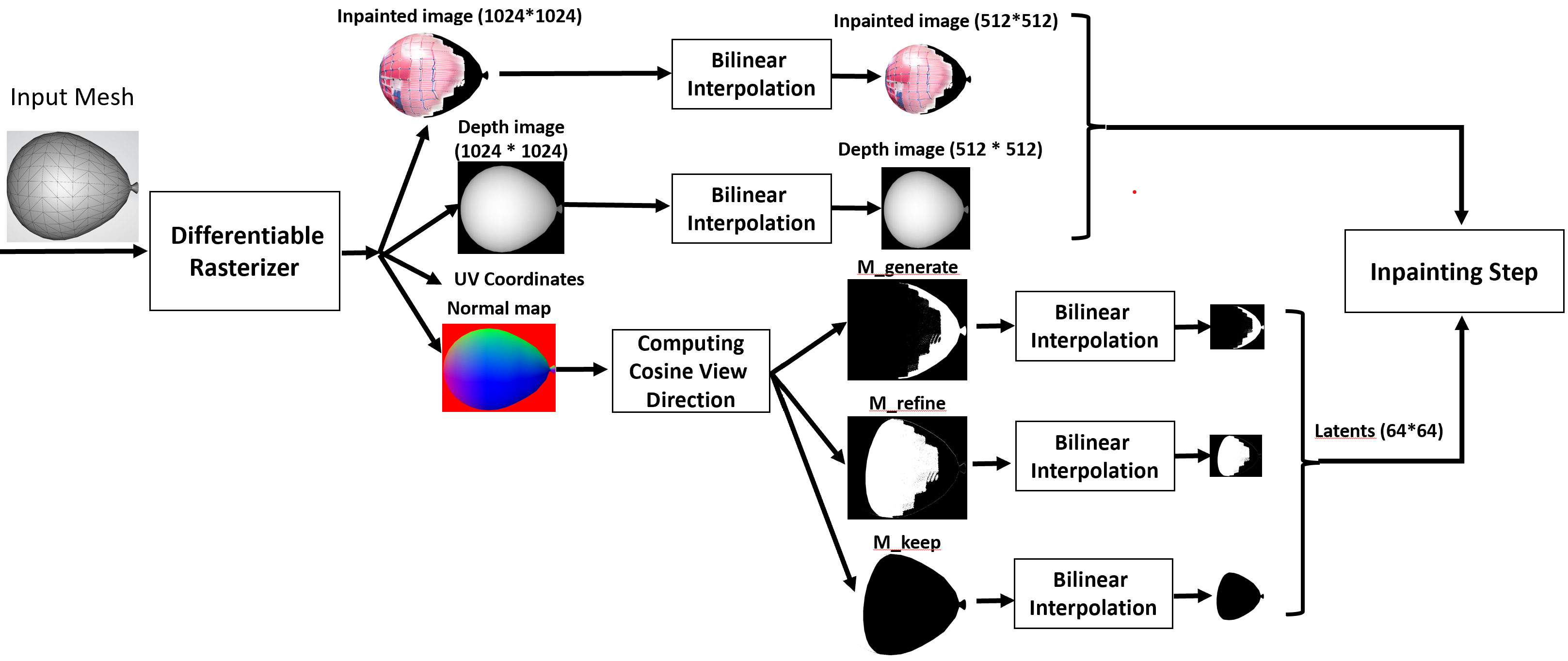}}
            \caption{Visualization of \textit{rendering} in the \textit{texture generation} step. For each camera viewpoint, we render the object using a differentiable renderer and extract rendering buffer including painted viewpoint image, depth maps, normal maps, and UV coordinates. Then, using the normal map, obtained from the differentiable renderer, we compute the \textit{view direction cosine}, and then generate three regions (masks), explained above, for each viewpoint: $M_{generate}, M_{refine}$ and $M_{keep}$. These three regions will serve as input to the next step of the texturing process to enforce the consistency in the output texture.}
            \label{DifferentiableRenderingSteps}
            \end{center}
            \vskip -0.1in
    \end{figure}

    \vskip -0.07in
    Due to the stochastic nature of the diffusion models, applying the iterative painting process for texturing process, explained in \cref{TextureGenerationStep}, would leads to having inconsistent textures with noticeable seams on the appearance of the object. To resolve this issue, we take a dynamic partitioning approach similar to \cite{TEXTure, Text2Tex}, where each rendered viewpoint is divided to three regions based on a concept called \textit{view direction cosine}. The \textit{view direction cosine} is utilized to determine how the surface is facing the viewer. Mathematically, the view direction cosine is represented as the cosine of the angle between two vectors: the \textit{view direction vector} which is the direction of the viewer's (in our case, the camera) line of sight, and \textit{mesh's surface normal vectors} associated with each face or vertex of the mesh. These normals indicate the direction perpendicular to the surface at that point (vertex):
         
     \vskip -0.22in
     \begin{equation}
        \text { viewcos }=\cos (\phi)=\frac{\vec{{V}} . \vec{N}}{ |\vec{V}| \cdot |\vec{N}|}
    \end{equation}
    \vskip -0.18in

    where $\vec{V}$ is the view vector, a vector starting from the camera position points toward a point on the surface of the 3D mesh object, $\vec{N}$ is the surface normal, vector perpendicular to the mesh surface, and $\phi$ is the angle between the view vector and surface normal. During the rendering process, this helps determine how to project and render 3D objects onto a 2D screen, ensuring that objects are viewed from the correct perspective. More specifically, the \textit{view direction cosine} is computed for each rendered viewpoint, and the rendered viewpoint is partitioned into three regions based on the value of \textit{view direction cosine}: 
     1) \textit{generate} - the regions have been viewed for the first time by the camera and have never been painted during the texturing process, 2) \textit{refine} - the regions that have already been viewed and painted from a viewpoint in previous iterations, but is now seen from a better camera angle (i.e. higher \textit{view direction cosine} values) and should be painted again, and 3) \textit{keep} - the regions that have already been viewed and painted from a good camera angle and should not be painted again. \cref{RemarkMathematicTexGen} presents further details and the differentiable mathematical representation of our \textit{texture generation} step.

    \subsection{Mathematical Representation}
    {
        \label{RemarkMathematicTexGen}
        Mathematically, the texture-generation process can be formulated as steps below (see \cref{TextureGenerationSteps}): (i) \textit{rendering}, render the object using a differentiable renderer \cite{NvidiffRast} and extract a rendering buffer $h^i$ corresponding to the camera viewpoint $i$, including painted viewpoint image, depth maps, normal maps, and UV coordinates:
    
        \begin{equation}
           h^i = f_{render}(a, v_{gen})
        \end{equation}
    
        where $a$ is the 3D asset’s parameters including texture image, the \textit{view direction cosine} for the current view, \textit{view direction cosine} for previous iterations, $v_{gen}$ is the camera viewpoints for texture-generation process, and $i \in v_{gen}$ is the current viewpoint. Using the normal map, obtained from the differentiable renderer, we then compute the \textit{view direction cosine}, and then generate three regions (masks), explained above, for each viewpoint: $M_{generate}, M_{refine}$ and $M_{keep}$. These three regions will serve as input to the next step of the texturing process to enforce the consistency in the output texture. \cref{DifferentiableRenderingSteps} visually demonstrates the details of the output generated by the differentiable renderer. (ii) \textit{Depth-aware painting}, at each denoising step $t$, the depth-aware diffusion process can be formulated as below:
        
        \vskip -0.2in
        \begin{equation}
            \begin{split}    
               y^i_{t-1} &=  f_{controlnet}(h^i, t, c) \\
               x^i_{t-1} &=  \mu(x^i_{t}, y^i_{t-1}, t, c) + \sigma_{t}z,  \;\;\; z \sim  \mathcal{N}(\mathbf{0}, \mathbf{I})
            \end{split}
        \end{equation}
    
        where $f_{controlnet}$ represents the ControlNet model whose parameters are fixed during the training, $h^i$ is the differentiable renderer output, $y^i_{t-1}$ is the control signal generated by the ControlNet model at timestep $t$ for viewpoint $i$, and $c$ is the prompt embedding obtained from the CLIP encoder model \cite{CLIP}. To enforce the consistency during the painting process, we follow \cite{BlendedLatentDiffusion} to incorporate latent mask blending in the latent space of the diffusion process. Specifically, by referring to the part that we wish to modify as $m_{blended}$ and to the remaining part as $m_{keep} = 1 - m_{blended}$, we blend the two parts in the latent space, as the diffusion progresses. This modifies the sampling process in a way that the diffusion model does not change the \textit{keep} regions of the rendered image. Therefore, the latent variable at the current diffusion timestep $t$ is computed as:
    
        \vskip -0.2in
        \begin{equation}
            z^i_t \leftarrow z^i_t \odot m_{\text {blended}}+z^i_{O_t} \odot\left(1-m_{\text {blended}}\right)
        \end{equation}
    
        where $z^i_{O_t}$ is the latent code of the original rendered image with added noise at timestep $t$ and for the viewpoint $i$ and $m_{blended}$ is the painting mask defined below:
    
        \vskip -0.2in
        \begin{equation}
            m_{blended}= \begin{cases}M_{\text {generate }}, & t \leq(1-\alpha) \times T \\ M_{\text {generate }} \cup M_{\text {refine }}, & t>(1-\alpha) \times T\end{cases}
        \end{equation}
    
        where $T$ is the total number of diffusion denoising steps and $\alpha$ is the refining strength which controls the level of refinement over the viewpoints - the larger it is, the less strong refinement we will have during the multi-viewpoint texturing process. Lastly, when the blending latent diffusion process is done, we output the image by decoding the resultant latent using the pre-trained variational autoencoder (VAE) existing in the stable diffusion pipeline.
    }

    \subsection{Modifications To Make Texture Generation Step Differentiable}
    \label{ModificationsTextureGeneration}
    {
       \noindent \textbf{Differentiable Camera Pose Computation}. 
       {
           Instead of using traditional camera positioning, to enable end-to-end training with backpropagation through the camera positioning steps, we provide a re-implementation of this procedure using PyTorch \cite{Pytorch} operations, preserving gradient flow. More specifically, camera poses are computed using statically generated camera-to-world transformation matrices from spherical coordinates (elevation and azimuth). This process involved the following steps: (i) spherical to cartesian conversion - given azimuth $\phi$, elevation $\theta$, and radius $r$, the camera position $c \in \mathbb{R}^3$ is computed as:
           
           \vspace{-5pt}
           \begin{equation}
                \begin{aligned}
                & x=r \cos (\theta) \sin (\phi), \\
                & y=-r \sin (\theta), \\
                & z=r \cos (\theta) \cos (\phi), \\
                & \mathbf{c}=[x, y, z]^{\top}+\mathbf{t},
                \end{aligned}
            \end{equation}  
            \vspace{-5pt}
            
            where $\mathbf{t}$ is the target point.
            (ii) Look-at matrix construction - the rotation matrix $R \in \mathbb{R}^{3\times3}$ is computed using a right-handed coordinate system by constructing orthonormal basis vectors:
    
            \vspace{-5pt}
            \begin{equation}
                \begin{array}{lr}
                \mathbf{f}=\text { normalize }(\mathbf{c}-\mathbf{t}) & \text { (forward vector) }, \\
                \mathbf{r}=\text { normalize }(\mathbf{u p} \times \mathbf{f}) & \text { (right vector) }, \\
                \mathbf{u}=\text { normalize }(\mathbf{f} \times \mathbf{r}) & \text { (up vector) },
                \end{array}
            \end{equation}
            \vspace{-5pt}
    
            where $\mathbf{up}=[0,1,0]^T$. The camera pose matrix $T \in \mathbb{R}^{4\times4}$ is then presented as:
    
            \vspace{-5pt}
            \begin{equation}
                \mathbf{T}=\left[\begin{array}{ll}
                \mathbf{R} & \mathbf{c} \\
                \mathbf{0}^{\top} & 1
                \end{array}\right]
            \end{equation}
            \vspace{-5pt}
    
            This modification transforms a previously non-trainable pre-processing step into a fully differentiable component, allowing camera positions and orientations to participate in gradient-based learning, particularly during texture preference learning where the texture is optimized to match geometry-aware reward signals.
        }

        \noindent \textbf{Differentiable Viewpoint Dynamic Partitioning (Masking).}
        {
            In the original implementation of the texture generation step \cite{InTex}, dynamic viewpoint partitioning for multi-view texturing was performed using hard, non-differentiable binary masks based on thresholding and bitwise operations. Specifically, three disjoint masks were used to guide different behaviors across image regions during training: (i) a generation mask $M_{generate}$, (ii) a refinement mask $M_{refine}$, and (iii) a keep mask $M_{keep}$. Let $c \in \mathbb{R}^{H \times W}$ denotes the count map (indicating coverage across viewpoints), and $v, v_{old} \in \mathbb{R}^{H\times W}$ be the view-consistency scores for the current and cached viewpoints, respectively. The original masks were defined as:

            \begin{equation}
                \begin{aligned}
                M_{\text {generate }} & =1[c<0.1] \\
                M_{\text {refine }} & =1\left[v>v_{\text {old }}\right] \cdot \neg M_{\text {generate }} \\
                M_{\text {keep }} & =\neg M_{\text {generate }} \cdot \neg M_{\text {refine }}
                \end{aligned}
            \end{equation}
            
            where $1[\cdot]$ denotes the indicator function and logical operators like, $\neg$ produce discrete masks, making them non-differentiable and unsuitable for gradient-based optimization. To enable end-to-end differentiability, we replace these hard decisions with soft, continuous approximations using the sigmoid function. Let $\sigma_k(x)=\frac{1}{1+\exp (-k x)}$ denote the sigmoid function with steepness $k$. We define soft masks as follows: (i) soft generation mask -

            \vspace{-5pt}
            \begin{equation}
                M_{\text {generate }}^{\text {soft }}=1-\sigma_k(c-0.1)
            \end{equation}
            
            This approaximates $1[c < 0.1]$, where $k=100$ ensures a sharp transition.(ii) soft refinement mask - 

            \vspace{-5pt}
            \begin{equation}
                M_{\text {refine }}^{\text {soft }}=\sigma_k\left(v-v_{\text {old }}\right) \cdot\left(1-M_{\text {generate }}^{\text {soft }}\right)
            \end{equation}
            
            This replaces the hard bitwise AND with elementwise multiplication and soft comparison. (iii) soft keep mask - 
            \begin{equation}
                M_{\text {keep }}^{\text {soft }}=\left(1-M_{\text {generate }}^{\text {soft }}\right) \cdot\left(1-M_{\text {refine }}^{\text {soft }}\right)
            \end{equation}
            \vspace{-10pt}
            
            These soft masks retain the semantics of the original binary partitioning but allow gradients to propagate through all mask operations, enabling joint optimization with the diffusion model during our texture preference learning step. We also apply bilinear interpolation in the zooming operation to maintain spatial smoothness across render resolutions.
        }
    }

}

\section{Remarks on Memory-saving Details in The Texture Preference Learning Step}
\label{RemarksTexturePreferenceLearning}
{
    \noindent \textbf{Activation Checkpointing.}
    {
        Gradient or activation checkpointing \cite{GradientCheckpointing} is a technique that trades compute for memory. Instead of keeping tensors needed for backward alive until they are used in gradient computation during backward, forward computation in checkpointed regions omits saving tensors for backward and recomputes them during the backward pass. In our framework, we apply two levels of gradient checkpointing to our pipeline. First, \textit{low-level checkpointing}, where we apply the checkpointing technique to the sampling process in the diffusion model by only storing the input latent for each denoising step, and re-compute the UNet activations during the backpropagation. Second, \textit{high-level checkpointing}, where we apply the checkpointing technique to every single-view texturing process in our multi-view texturing algorithm. Practically, what it means is that, for each viewpoint during the multi-view iterative texturing process, intermediate tensors inside the per-view texturing are not saved. Instead, only the inputs (camera pose, prompt embeddings, etc.) are saved. Then, during the backward pass, we recompute the forward pass of the per-view texturing again so gradients can flow backward. Overall, this technique allows to keep memory usage constant even across many viewpoints and diffusion steps which consequently leads to scaling to more views and using higher-resolution renders or more diffusion steps to have a more realistic and visually appealing texture image without any memory issues.
    }

    \noindent \textbf{Low-Rank Adaptation (LoRA).}
    {
        LoRA is a technique, originally introduced in \cite{LoRA} for large language models, that keeps the pre-trained model weights fixed and adds new trainable low-rank matrices into each layer of the neural architecture which highly reduces the number of trainable parameters during the training process. Mathematically, for a layer with base parameters $W$ whose forward pass is $h=Wx$, the LoRA adapted layer is $h=Wx + BAx$, where $A$ and $B$ are the low-rank trainable matrices and the $W$ remains unchanged during the training process. In our case, we apply the LoRA techniques to the U-Net architecture inside the latent diffusion model (LDM) \cite{SoftRasterizer} which lies at the heart of our texture generation step. This allows us to fine-tune the LDM using a parameter set that is smaller by several orders of magnitude, around 1000 times fewer, than the full set of LDM parameters.
    }

}

\section{Remarks on Principal Curvature Directions}
\label{RemarkPrincipalCurvature}
{        
    Principal curvature directions at a point on a surface describe how the surface bends in different directions. Curvature is defined as $k=\frac{1}{R}$ where $R$ is the radius of the circle that best fits the curve in a given direction. Among all possible directions, two orthogonal directions, called the principal curvature directions, are of special interest: the maximum and minimum curvature directions. The minimum curvature direction corresponds to the direction in which the surface bends the least (i.e., the largest fitting circle), while the maximum curvature direction corresponds to the direction in which it bends the most (i.e., the smallest fitting circle). See \cref{PrincipalCurvatures} for an illustration. Additionally, the mean curvature at a point is defined as the average of the two principal curvatures: $H=\frac{1}{2}(k_1+k_2)$. We use the Libigl library~\cite{libigl} to compute both the mean curvature values and the principal curvature directions on our 3D mesh surfaces.

    \vskip -0.07in    
    \begin{figure}[ht]
        \vskip 0.2in
        \begin{center}
        \centerline{\includegraphics[scale=0.2]{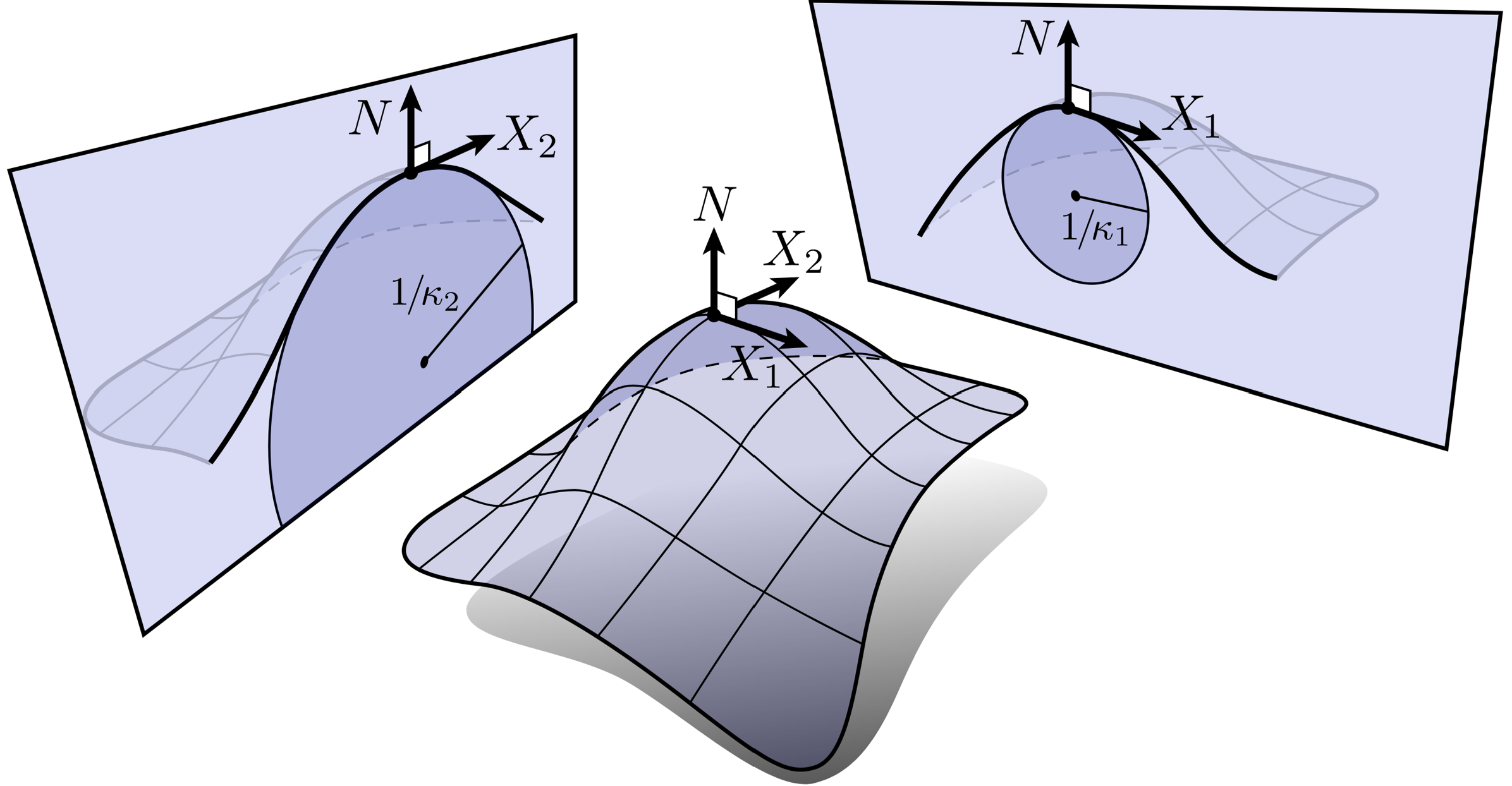}}
        \caption{Visualization of principal curvatures (Image Credit: \href{http://wordpress.discretization.de/geometryprocessingandapplicationsws19/a-quick-and-dirty-introduction-to-the-curvature-of-surfaces/}{Keenan Crane}). The minimum curvature direction corresponds to the direction in which the surface bends the least (i.e., the largest fitting circle), while the maximum curvature direction corresponds to the direction in which it bends the most (i.e., the smallest fitting circle).}
        \label{PrincipalCurvatures}
        \end{center}
        \vskip -0.2in
    \end{figure}
}

    \section{Emergence of Repetitive Patterns through Geometry-Aligned Sampling.}
    \label{EmergenceRepetitivePatterns}
    {
        An interesting observation in our results (see  \cref{CurvTexAlignQualitativeResultShort} and \cref{CurvTexAlign_Qualitative}) is the emergence of repetitive texture patterns after fine-tuning with the geometry-texture alignment reward. This behavior stems from the mismatch between the number of curvature vectors and the number of texture gradients. Specifically, curvature is a vertex-level attribute defined per mesh vertex, whereas texture gradients are defined per pixel in the 2D UV space. For example, in the bunny mesh example, we have 4487 curvature vectors (one per vertex), but over one million texture gradients from a $1024 \times 1024$ texture image. To compute cosine similarity between curvature vectors and texture gradients, both must be in the same space and of the same dimension. We address this by sampling the texture gradients at the UV coordinates corresponding to mesh vertices, the same positions where curvature vectors are defined. This results in two aligned tensors (both of size 4487 in the bunny case), enabling the reward function to compare directional alignment directly. Over time, this differentiable sampling mechanism guides the model to place strong texture features (e.g., edges) at those specific UV coordinates. As these positions remain fixed (being tied to the mesh vertices), the model learns to reinforce edge features at the same UV locations across views and iterations. This alignment results in visually repetitive patterns in the generated textures, especially in regions of high curvature.

    }

\section{Remarks on Texture Features Emphasis Reward}
\label{RemarksTextureFeaturesEmphasisReward}
{
    This reward encourages texture patterns that emphasize 3D surface structure while maintaining perceptual richness through color variation. It consists of two components: (i) \textit{texture-curvature magnitude alignment} which strengthens texture variations in high-curvature regions to enhance capturing geometric structures, and (ii) \textit{colorfulness} inspired by \cite{MultiColor, ColorfulnessReward}, which discourages desaturated textures by encouraging vibrant color use, measured through the standard deviation and mean distance of opponent color channels from neutral gray. Let $T(x,y) \in [0,1]^3$ denote the RGB texture at pixel $(x,y)$ and $C(x,y) \in [0,1]$ be the normalized curvature map in UV space. The \textit{texture-curvature alignment} term is mathematically defined as the negative mean squared error (MSE):
    
    \vskip -0.18in
    \begin{equation}
        R_{magnitude}=-\frac{1}{N} \sum_{x, y}({\nabla} {\mathbf{T}}(x, y)-{C}(x, y))^2
    \end{equation}
    \vskip -0.1in
    
    where $\nabla \mathrm{T}(x, y)=\sqrt{\left\|\frac{\partial \mathrm{T}}{\partial x}(x, y)\right\|^2+\left\|\frac{\partial \mathrm{T}}{\partial y}(x, y)\right\|^2}$ is the texture gradient magnitude and $N$ is the total number of pixels in the texture image.

    Then, we augment this reward with a colorfulness term based on opponent color channels~\cite{ColorfulnessReward}. Let \( R(x,y), G(x,y), B(x,y) \in \mathbb{R}^{H \times W} \) denote the red, green, and blue channels of the texture. We compute the opponent axes:

    \vspace{-10pt}
    \begin{equation}
    rg = R - G, \quad yb = \frac{1}{2}(R + G) - B
    \end{equation}
    \vspace{-10pt}
    
    the \textit{colorfulness} term is then defined as:
    
    \vskip -0.05in
    \begin{equation}
        \label{Colorfulness_reward}
        R_{color}=\sigma_{r g}+\sigma_{y b}+0.3\left(\left|\mu_{r g}\right|+\left|\mu_{y b}\right|\right)
    \end{equation}                
    
    where $\sigma_{rg}$, $\sigma_{yb}$, $\mu_{rg}$, and $\mu_{yb}$ are the standard deviations and mean colors offsets corresponding to color opponent components $rg(x,y) = R(x,y) - G(x,y)$ and $yb(x,y) = \frac{R(x,y)+G(x,y)}{2} - B(x,y)$, respectively. 
    The full \textit{texture features emphasis} reward combines both components:
    
    \vskip -0.1in
    \begin{equation}
        R_3 = \alpha_m R_{magnitude} + \alpha_c R_{color}
    \end{equation}                
    
    where $\alpha_m$ and $\alpha_c$ are weights for the \textit{texture-curvature magnitude alignment} and \textit{colorfulness} terms, respectively.
}

\section{Remarks on Geometry-Aware Texture Colorization}
\label{GeometryGuidedTextureColorizationReward}
{    
    \subsection{Geometry-Aware Texture Colorization Reward Design}    
    {
        To encode surface curvature into texture color, we design a reward function that encourages warm colors (e.g., red, yellow) in regions of high curvature and cool colors (e.g., blue, green) in regions of low curvature. Given a per-pixel scalar curvature map $C(x,y) \in [-1,1]$, and a curvature threshold $T \in [-1,1]$, let $I_r(x,y)$ and $I_b(x,y)$ denote the red and blue channels of the predicted texture at pixel $(x,y)$, respectively. For pixels where $C(x,y) > T$, we encourage warm colors by rewarding a higher red-blue channel difference $\Delta_{rb}(x,y)=I_r(x,y)-I_b(x,y)$. Conversely, for $C(x,y) < T$, we encourage cool colors by penalizing this difference. The final reward is the average $\Delta_{rb}$ across all pixels:

        \vskip -0.3in
        \begin{equation}
            \label{colroization_reward}
            R_2=\frac{1}{N} \sum_{x, y} \begin{cases}\Delta_{r b}(x, y), & \text { if } C(x, y)>T \\ -\Delta_{r b}(x, y), & \text { if } C(x, y) \leq T\end{cases}
        \end{equation}
        \vskip -0.11in
    
        where $N$ is the total number of pixels. This formulation captures the desired color-geometry correlation. However, its hard conditional logic makes it non-differentiable and unsuitable for gradient-based training. To resolve this, we replace hard thresholds with smooth, sigmoid-based approximations, enabling end-to-end differentiability.
    
        Note that to compute the curvature-quided colorization reward, we need the mean curvature 2D map as a way to incorporate mesh's geometry information into texture learning objectives (\cref{colroization_reward}). To this end, we construct a mean curvature 2D map, a 2D image in UV space that encodes the mean curvature of the underlying 3D mesh surface and then compute the rewards in the 2D UV space by comparing the texure image and the 2D curvature map. This process involves the following key steps.
        
        \noindent \textbf{UV-Pixel mapping.} Each mesh vertex is associated with a UV coordinate $(u,v) \in [0,1]^2$. These UV coordinates are scaled to a discrete $H \times W$ pixel grid representing the output texture space. 
        
        \noindent \textbf{Per-face barycentric interpolation.} For each triangular face of the mesh, we project its UV coordinates to the texture grid, forming a triangle in 2D space. We then compute a barycentric coordinate transform that expresses any point within the triangle as a convex combination of its three corners. 
        
        \noindent \textbf{Compute barycentric coordinates.} We iterate over each pixel inside the UV triangle's bounding box and compute its barycentric coordinates. If a pixel lies within the triangle, its mean curvature value is interpolated from the curvature values at the triangle's corresponding 3D vertices using:
    
        \vspace{-15pt}
        \begin{align}
        \operatorname{val} =\ &\operatorname{bary}[0] \cdot \operatorname{curv}_v[\operatorname{tri\_v}[0]] + \nonumber \\
                             &\operatorname{bary}[1] \cdot \operatorname{curv}_v[\operatorname{tri\_v}[1]] + \nonumber \\
                             &\operatorname{bary}[2] \cdot \operatorname{curv}_v[\operatorname{tri\_v}[2]]
        \end{align}
        \vspace{-10pt}
        
        where $\operatorname{bary}[i]$ is the $i$-th barycentric coordinate of the pixel relative to the UV triangle, $\operatorname{tri\_v[i]}$ is the index of the $i$-th 3D vertex corresponding to the UV triangle corner, and $\operatorname{curv}_v$ is the mean curvature array, containing one scalar curvature per 3D vertex. This produces a smoothly interpolated curvature value for each pixel within the UV triangle. 
        
        \noindent \textbf{Accumulation and averaging.} Each pixel may be covered by multiple UV triangles. Therefore, we accumulate curvature values and record the number of contributions per pixel. After processing all triangles, we normalize the curvature value at each pixel by the number of contributions, producing a final per-pixel average curvature. The output is then a dense 2D curvature texture defined over the UV domain. This texture captures the surface's geometric structure and is suitable for use in differentiable training objectives or geometric regularization.
    
        \vspace{-5pt}
    }

    \subsection{Geometry-Aware Texture Colorization Experiment and Results}    
    {
        This task aims to colorize textures based on surface bending intensity, represented by mean curvature, an average of the minimal and maximal curvature directions, on the 3D mesh. Specifically, the model is encouraged to apply warm colors (e.g., red, yellow) in high-curvature regions and cool colors (e.g., blue, green) in low-curvature areas. \cref{GeometryGuidedTextureColorizationQualitativeResults} shows qualitative results on rabbit and cow mesh objects colorized with different textual prompts. As illustrated, our method consistently adapts texture colors, regardless of initial patterns, according to local curvature and successfully maps warmth and coolness to geometric variation (more results in \cref{GeometryGuidedTextureColorizationQualitativeResults}).
    }

}

\section{Attempts to Overcome Reward Hacking}
\label{AttemptsToOvercomeRewardHacking}
{
    To prevent the fine-tuned diffusion model from diverging too far from the original pre-trained model and to mitigate issues such as overfitting and reward hacking, we incorporate three regularization strategies: KL divergence \cite{DPOK}, LoRA scaling \cite{LoRA_Scaling}, and a colorfulness incentive \cite{ColorfulnessReward}. These techniques are added as regularization terms to each reward function we design and are evaluated for their effectiveness during training. While KL regularization offers a principled approach to reducing reward overfitting, we find that LoRA scaling and the colorfulness incentive \cite{ColorfulnessReward} consistently yield better empirical results. We detail each of these techniques in the following.
    
    \noindent \textbf{KL-Divergence.} Inspired by \cite{Draft} and similar to \cite{DPOK}, we introduce a regularization penalty that encourages the fine-tuned model to stay close to the behavior of the pre-trained Stable Diffusion model. Specifically, we penalize the squared difference between the noise predictions of the pre-trained and fine-tuned U-Net models at the final denoising step. Let $\epsilon_{\mathrm{pt}}$ and $\epsilon_{\mathrm{ft}}$ denote the predicted noise from the pre-trained and fine-tuned models, respectively. We define the penalty term as:
    
    \begin{equation}
        \beta_{\mathrm{KL}}\left\|\epsilon_{\mathrm{ft}}-\epsilon_{\mathrm{pt}}\right\|_2^2
    \end{equation}
    
    and add it to the overall reward during training. The coefficient $\beta_{\mathrm{KL}}$ controls the strength of this regularization and is varied across different experimental settings. Intuitively, this penalty discourages the fine-tuned model from diverging too far from its pre-trained counterpart, helping to mitigate issues such as catastrophic forgetting or overly narrow specialization. This approach is justified by the fact that, under fixed variance, the KL divergence between two Gaussian noise distributions simplifies to the squared difference between their means. In the context of conditioned diffusion models, noise prediction $\epsilon$ is conditioned on a noised latent $x_t$. We compute this penalty at the final denoising timestep $t = 1$, just before image reconstruction. For each training batch, we compute $\epsilon_{\mathrm{pt}}$ by forwarding the frozen pre-trained model (with gradients disabled), and $\epsilon_{\mathrm{ft}}$ by forwarding the fine-tuned model. The squared difference, scaled by $\beta_{\mathrm{KL}}$, is then added to the reward signal.

    \noindent \textbf{LoRA scaling.} Inspired by \cite{Draft} and similar to \cite{LoRA_Scaling}, we scale the LoRA parameters by a factor in the range $(0, 1]$ to modulate their effect during fine-tuning. Empirically, we observe that factors close to 1, specifically within $[0.9, 1]$, consistently yield the best performance in our experiments.
    
    \noindent \textbf{Colorfulness incentive.} Inspired by \cite{MultiColor, ColorfulnessReward}, we discourage desaturated texture images by encouraging vibrant color use, measured through the standard deviation and mean distance of opponent color channels from neutral gray. To this end, we augment our reward with a colorfulness term based on opponent color channels~\cite{ColorfulnessReward}. Let \( R(x,y), G(x,y), B(x,y) \in \mathbb{R}^{H \times W} \) denote the red, green, and blue channels of the texture. We compute the opponent axes:

    \vspace{-15pt}
    \begin{equation}
    rg = R - G, \quad yb = \frac{1}{2}(R + G) - B
    \end{equation}
    \vspace{-10pt}
    
    the \textit{colorfulness} term is then defined as:
    
    \vskip -0.05in
    \begin{equation}
        \label{Colorfulness_reward}
        R_{color}=\sigma_{r g}+\sigma_{y b}+0.3\left(\left|\mu_{r g}\right|+\left|\mu_{y b}\right|\right)
    \end{equation}                
    
    where $\sigma_{rg}$, $\sigma_{yb}$, $\mu_{rg}$, and $\mu_{yb}$ are the standard deviations and mean colors offsets corresponding to color opponent components $rg(x,y) = R(x,y) - G(x,y)$ and $yb(x,y) = \frac{R(x,y)+G(x,y)}{2} - B(x,y)$, respectively. This term is incorporated into both our \textit{texture feature emphasis} and \textit{symmetry-aware texture generation} reward functions to promote visually rich and saturated outputs.

}

\section{Ablation Studies}
\label{Ablation}
{
    We include a focused ablation study to quantify the effect of regularization choices, LoRA scaling, and the memory/computation trade-off of our proposed configuration. Results and qualitative examples are shown in \cref{fig:ablation_regularizations} (regularization ablations) and \cref{fig:ablation_lora_all} (LoRA scaling ablations), and a compact memory/throughput comparison is reported in \cref{tab:ablation_memory}.

    \vspace{-15pt}

    \paragraph{Full vs. LoRA+Checkpointing fine-tuning.}
    {
        \cref{tab:ablation_memory} summarizes the memory and throughput trade-offs. Using LoRA in combination with activation checkpointing reduces the peak GPU memory substantially (from an infeasible full-fine-tuning footprint to a modest working set) and reduces the number of trainable parameters by several orders of magnitude, while maintaining acceptable throughput. The table demonstrates that our default configuration provides a realistic balance between memory cost and training performance for high-resolution texture adaptation. Therefore, when training on limited GPU memory, it is recommended to enable activation checkpointing and LoRA to dramatically reduce peak memory, as shown in \cref{tab:ablation_memory}.
    }    
    \vspace{-15pt}
    
    \paragraph{Fine-tuning with/without regularization strategies.}
    {
        \cref{fig:ablation_regularizations} compares two regularizers that we found important in practice: a colorfulness regularizer (\cref{fig:ablation_colorfulness}) and a LoRA-based regularization term that penalizes LoRA parameters changes (\cref{fig:ablation_lora_regularization}). Each subfigure shows three rows for the same example mesh: the input (before fine-tuning), the output after fine-tuning \emph{without} the regularizer, and the output after fine-tuning \emph{with} the regularizer. The colorfulness regularizer reduces undesirable saturations and preserves a more balanced color distribution across the texture image, which is visible in the bunny texture (\cref{fig:ablation_colorfulness}). The LoRA regularization term helps keep the fine-tuned model’s distribution close to the pre-trained one while allowing parameter updates to satisfy the reward objective. This yields texture images that both exhibit natural patterns and meet specific objectives (e.g., symmetry patterns in \cref{fig:ablation_lora_regularization}. Qualitatively, both regularizers improve perceptual texture quality compared to training without them, though the more beneficial choice depends on the reward and desired visual characteristics.
    }

    \vspace{-15pt}
    
    \paragraph{LoRA scaling.}
    {
        \cref{fig:ablation_lora_all} reports reward curves for multiple LoRA scale values under two reward objectives: symmetry-aware (\cref{fig:ablation_lora_symm}) and aesthetic (\cref{fig:ablation_lora_aes}). The plots show smoothed rewards over training iterations for each LoRA scale value. Across both objectives we observe a trade-off: larger LoRA scales accelerate early gains (they provide more expressiveness and thus faster improvement on the target reward), but they can also increase variance and, in some settings, cause less stable long-term behaviour. Smaller scales train more conservatively and tend to be more stable, but require more iterations to reach the same reward level. The two rewards behave slightly differently: the symmetry reward benefits noticeably from higher early expressiveness, while the aesthetic reward shows a more gradual, steady improvement for moderate scales. Based on these results, we recommend a moderate-to-high default (e.g., 0.7–0.9) and advise practitioners to tune upward only when faster short-term gains are needed and the risk of added variance is acceptable.
    }

    \begin{table}[ht]
      \centering      
      \caption{Ablation study comparing LoRA+checkpointing fine-tuning with full fine-tuning. Results show a substantial reduction in memory usage and trainable parameters when using LoRA with activation checkpointing. (OOM = Out of Memory)}
      \vspace{-5pt}
      \label{tab:ablation_memory}
      \scriptsize
      \setlength{\tabcolsep}{4pt} 
      \resizebox{\columnwidth}{!}{%
        \begin{tabular}{@{}lccc@{}}
          \toprule
          Configuration & Peak GPU mem (GB) & \#Params (M) & Throughput (sec/iter) \\
          \midrule
          Full fine-tuning                 & 300 (OOM) & 859.5   & OOM \\
          Ours (LoRA + Checkpointing)     & 20        & 0.797   & 40  \\
          \bottomrule
        \end{tabular}%
      }
    \end{table}

    \vspace{-10pt}

    \begin{figure}[ht]
      \centering
    
      \begin{subfigure}[t]{\linewidth}
        \centering
        \includegraphics[width=1\linewidth]{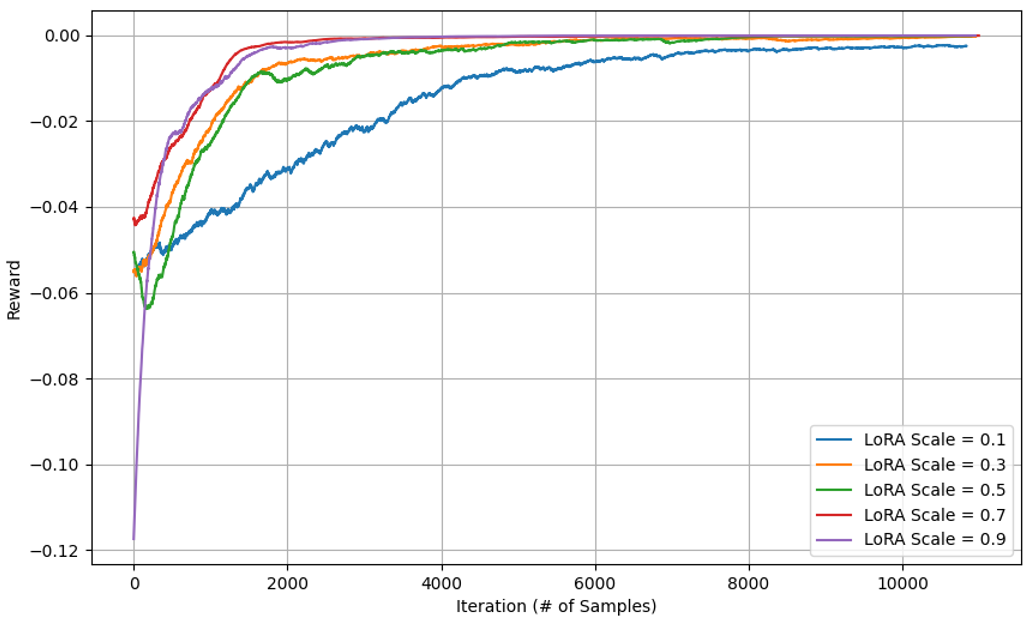}
        \caption{Ablation on LoRA scale measured with the symmetry-aware reward. Curves show smoothed reward versus training iterations for multiple LoRA scale values. Larger LoRA scales produce faster initial gains on the symmetry objective but can introduce higher variance; moderate-to-high scales achieve a good balance between speed and stability.}
        \label{fig:ablation_lora_symm}
      \end{subfigure}

      \begin{subfigure}[t]{\linewidth}
        \centering        
        \includegraphics[width=1.12\linewidth]{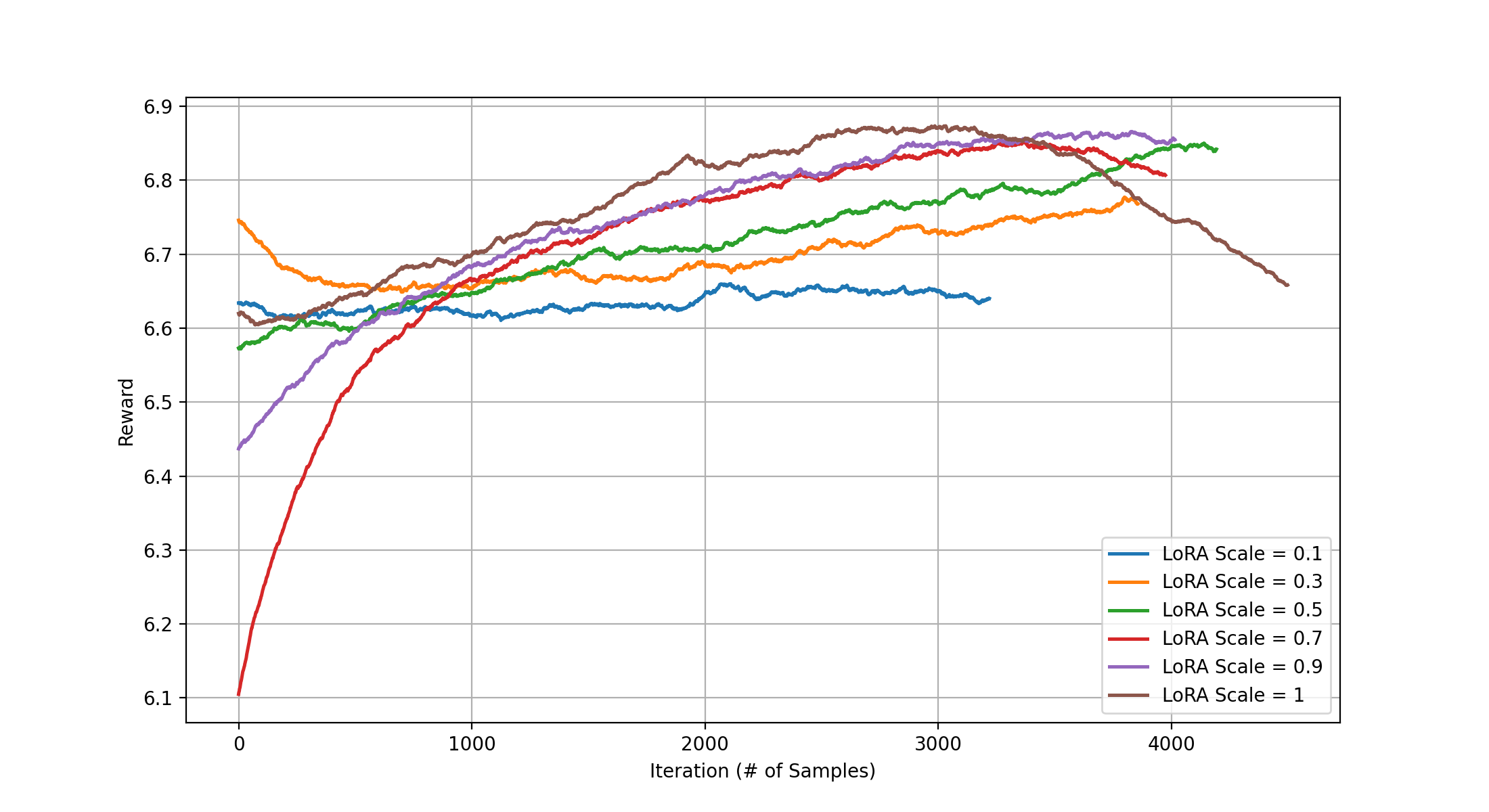}
        \caption{Ablation on LoRA scale measured with the aesthetic reward. Curves show the smoothed aesthetic reward over iterations for the same set of LoRA scales. The relative ranking of scales differs slightly from the symmetry objective, indicating that reward choice interacts with LoRA expressiveness.}
        \label{fig:ablation_lora_aes}
      \end{subfigure}
    
      \caption{Ablation study on different values of LoRA scale. (Top) symmetry-aware reward. (Bottom) aesthetic reward. In both plots each series corresponds to a different LoRA scale (see legends); curves were smoothed for visual clarity.}
      \label{fig:ablation_lora_all}
    \end{figure}

    \begin{figure}[ht]
      \centering
    
      \begin{subfigure}[t]{\linewidth}
        \centering
        \includegraphics[width=1.05\linewidth]{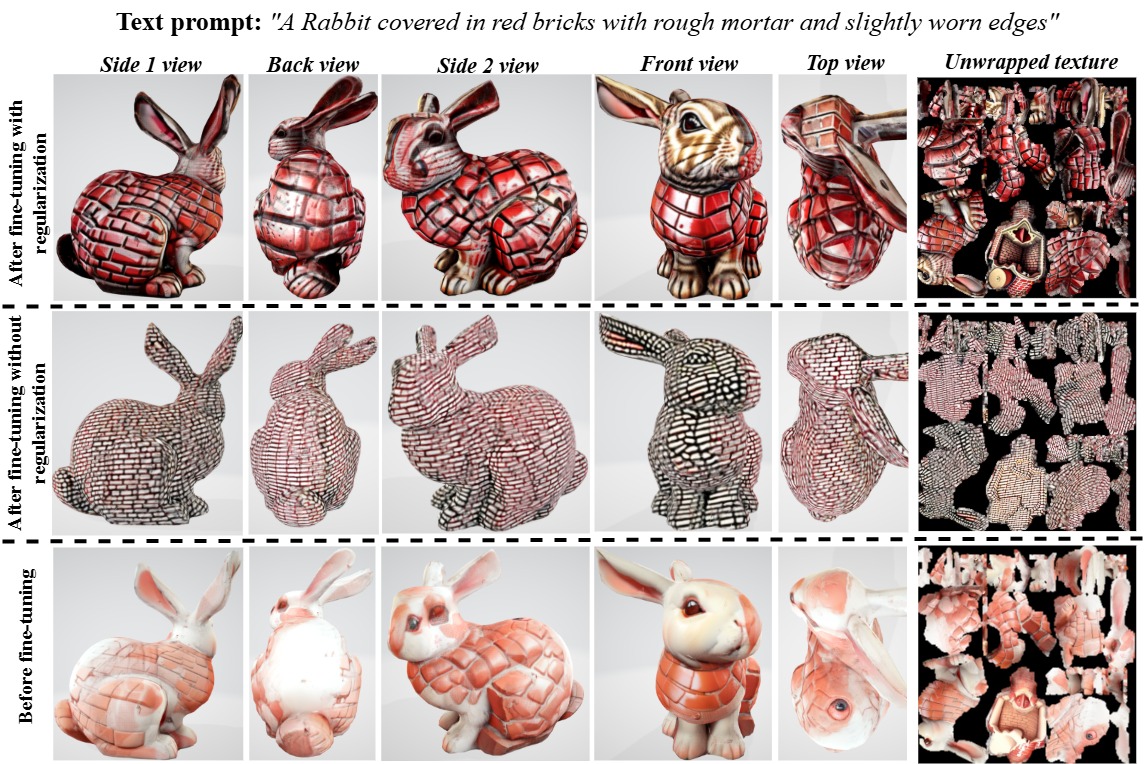}
        \caption{Ablation on colorfulness regularization. Each row shows (top to bottom): (1) result after fine-tuning with colorfulness regularization, (2) result after fine-tuning without colorfulness regularization, and (3) the input / before fine-tuning baseline. Columns show a set of canonical views (side / back / side2 / front / top) and the unwrapped texture. The colorfulness regularizer reduces excessive saturation and preserves perceptually balanced color distributions in the texture image while keeping geometric detail.}
        \label{fig:ablation_colorfulness}
      \end{subfigure}

        \vspace{6mm}
    
      \begin{subfigure}[t]{\linewidth}
        \centering
        \includegraphics[width=1.05\linewidth]{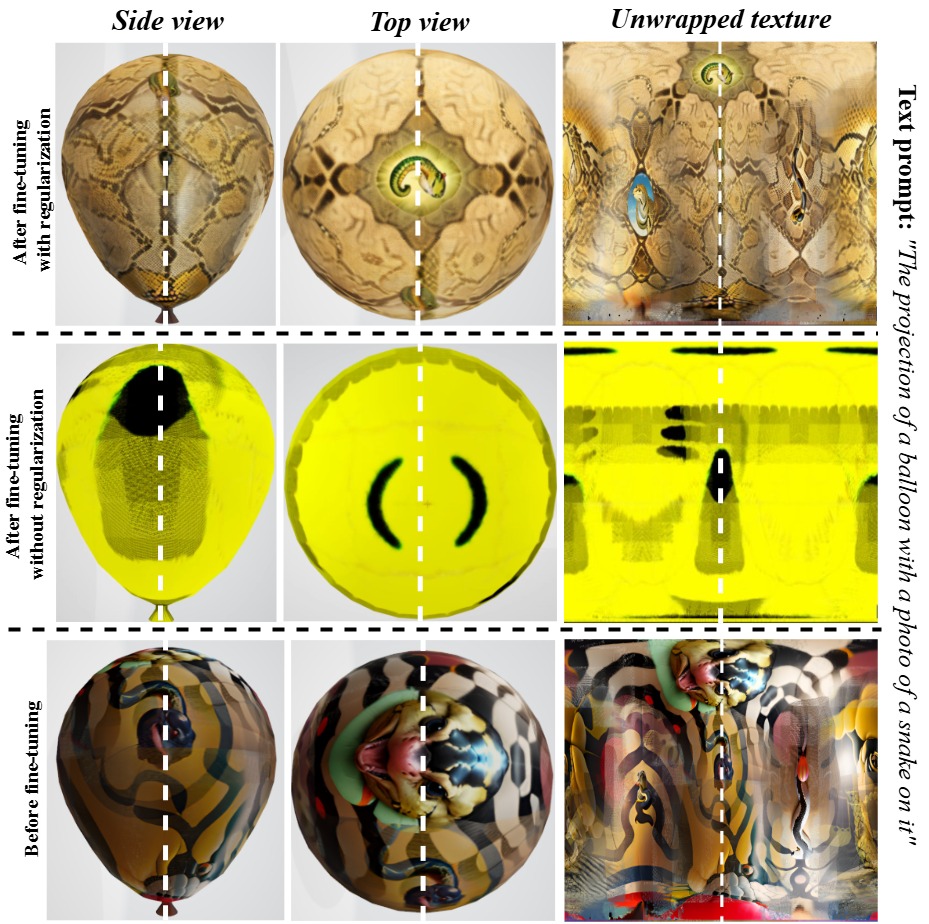}
        \caption{Ablation on LoRA scaling regularization. Rows are arranged as in \cref{fig:ablation_colorfulness}: (1) after fine-tuning with LoRA regularization, (2) after fine-tuning without LoRA regularization, and (3) before fine-tuning. Columns show side, top and the unwrapped UV texture. Using a mild-to-high LoRA regularization term helps keep the fine-tuned model’s distribution close to the pre-trained one while allowing parameter updates to satisfy the reward objective. This yields texture images that both exhibit natural patterns and meet specific objectives (e.g. symmetry patterns here).}
        \label{fig:ablation_lora_regularization}
      \end{subfigure}

        \vspace{3mm}
        
      \caption{Ablation studies on regularization strategies. (Top) colorfulness regularization; (Bottom) LoRA sacling regularization. Each subfigure compares the same mesh and texture image before fine-tuning, after fine-tuning without the regularizer, and after fine-tuning with the regularizer (see subcaptions for details).}    
      \label{fig:ablation_regularizations}
    \end{figure}

}

\section{Additional Experiments and Results}
\label{AdditionalExperimentsResults}
{   

    \begin{table}[t]
    \caption{A quantitative comparison on the bunny model, using various texture patterns, evaluates our fine‑tuned method against InTeX \cite{InTex} (before fine‑tuning) for geometry–texture alignment. Each column reports the number of aligned vector pairs (i.e., minimum curvature directions versus texture gradients) in UV space. Our approach yields significantly higher alignment counts, demonstrating that the generated textures respect the underlying geometric structure, whereas the InTeX \cite{InTex} baseline fails to achieve such alignment.\vspace{-15pt}}
    \label{QuantitativeResults}
    \begin{center}
    \begin{small}
    \begin{sc}
    \scalebox{0.83}{\begin{tabular}{lcccr}
    \toprule
    Texture pattern & After fine-tune & Before fine-tune \\
    \midrule
    Tiger stripe    & \textbf{630/4487 (14.04\%)} & 379/4487 (8.45\%)\\
    Black \& white line    & \textbf{826/4487 (18.41\%)} & 372/4487 (8.29\%)\\
    Checkerboard     & \textbf{485/4487 (10.81\%)} & 321/4487 (7.15\%) \\
    Brick    & \textbf{477/4487 (10.63\%)} & 302/4487 (6.73\%) \\
    Feather  & \textbf{556/4487 (12.39\%)} & 343/4487 (7.64\%) \\
    Weave  & \textbf{472/4487 (10.52\%)} & 299/4487 (6.66\%) \\
    \bottomrule
    \end{tabular}}
    \end{sc}
    \end{small}
    \end{center}
    \vskip -0.in
    \end{table}

    We show expanded results from the main paper featuring more texture image and more viewpoints of different 3D objects. \cref{Symmetry_Qualitative}, \cref{CurvTexAlign_Qualitative}, \cref{GeometryGuidedTextureColorizationQualitativeResults}, and \cref{TextureFeatureEmphasis_Qualitative} shows the qualitative results of the symmetry-aware, geometry-texture alignment, geometry-guided texture colorization, and texture features emphasis experiments, respectively. Moreover, a quantitative comparison on the \textit{geometry-texture alignment} is presented in \cref{QuantitativeResults}, using various texture patterns on the bunny object. It evaluates our fine-tuned method against InTeX \cite{InTex} (before fine-tuning) for geometry–texture alignment. Our approach yields significantly higher alignment counts, demonstrating that the generated textures respect the underlying geometric structure, whereas the InTeX \cite{InTex} fails to achieve such alignment. 
    
    We also perform additional experiments to extend results on higher-poly, occluded, and complex mesh objects in \cref{ExtendedResults_ComplexObjects}. Since our method modifies only texture (not geometry), self-occlusion and mesh resolution do not directly affect the outputs.

   \begin{figure*}[h]
            \begin{center}
            \centerline{\includegraphics[scale=0.137]{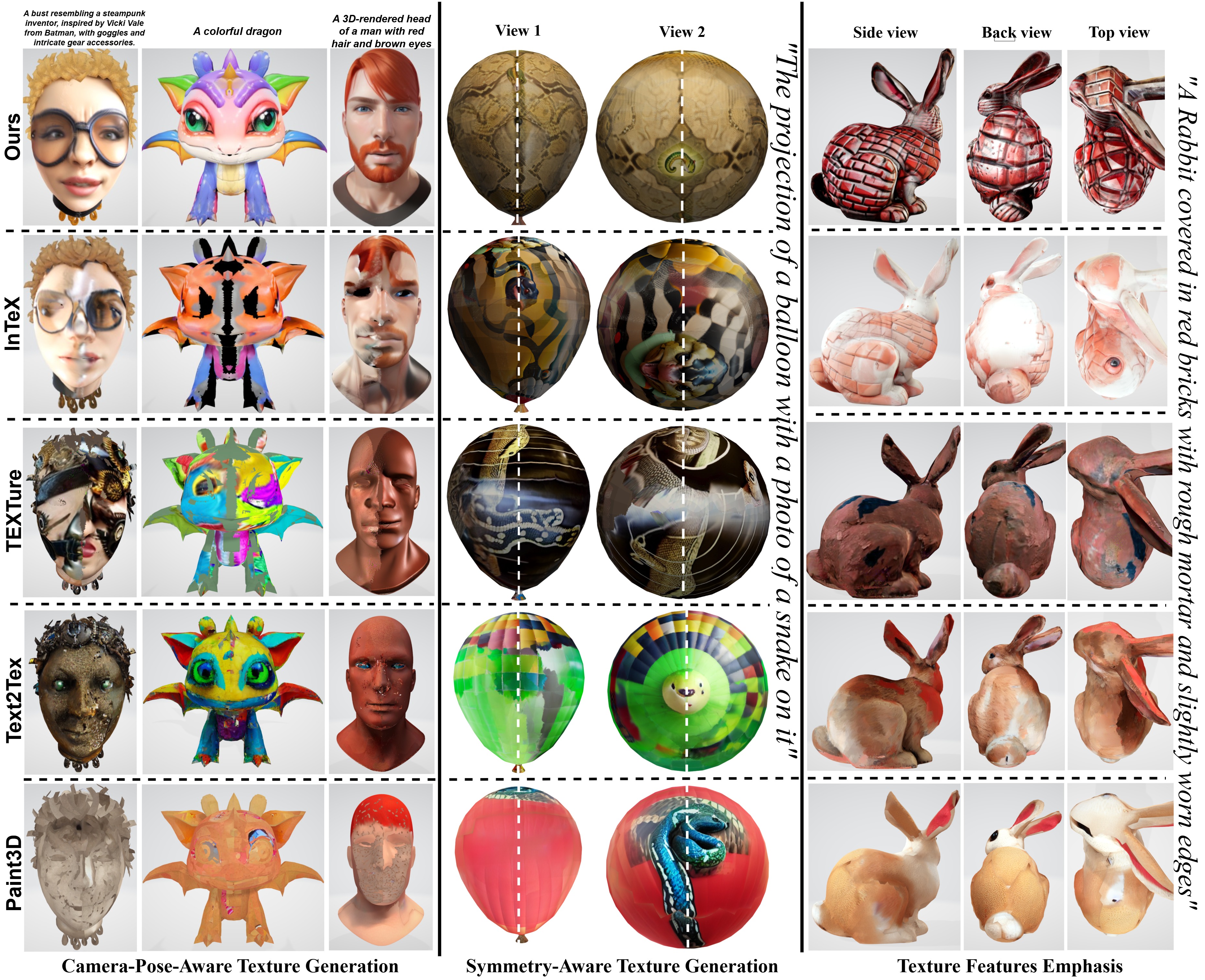}}
            \caption{\textbf{Left:} Qualitative results of the camera-pose-aware texture generation experiment on three different 3D mesh objects (female head, dragon, and male head). Our method produces textures with more aesthetic patterns that are more realistic and align more closely with the mesh’s semantic parts. \textbf{Middle:} Qualitative results for the symmetry‑aware experiment on a 3D balloon mesh. For each viewpoint, we present the rendered mesh with a vertical dashed line indicating the symmetry axis. Compared to InTeX \cite{InTex}, TEXTure \cite{TEXTure}, Text2Tex \cite{Text2Tex}, and Paint3D \cite{Paint3D}, our method produces textures that exhibit consistent patterns across symmetric regions of the mesh, confirming the reward’s ability to enforce symmetry. In contrast, without symmetry supervision, the textures on opposite sides often diverge significantly. \textbf{Right:} Qualitative results of the texture features emphasis experiment on a rabbit (bunny) object. We show the rendered 3D bunny from multiple viewpoints. Our method enhances texture features, such as edges and mortar, in proportion to mean curvature, a capability that all four baselines lack, often resulting in pattern-less (white) areas, particularly on the back and head of the rabbit. \vspace{-35pt}}          
            \label{ThreeRewards_Results}
            \end{center}            
    \end{figure*}

    \vskip -0.07in    
    \begin{figure*}[ht]
        \vskip 0.2in
        \begin{center}
        \centerline{\includegraphics[scale=0.27]{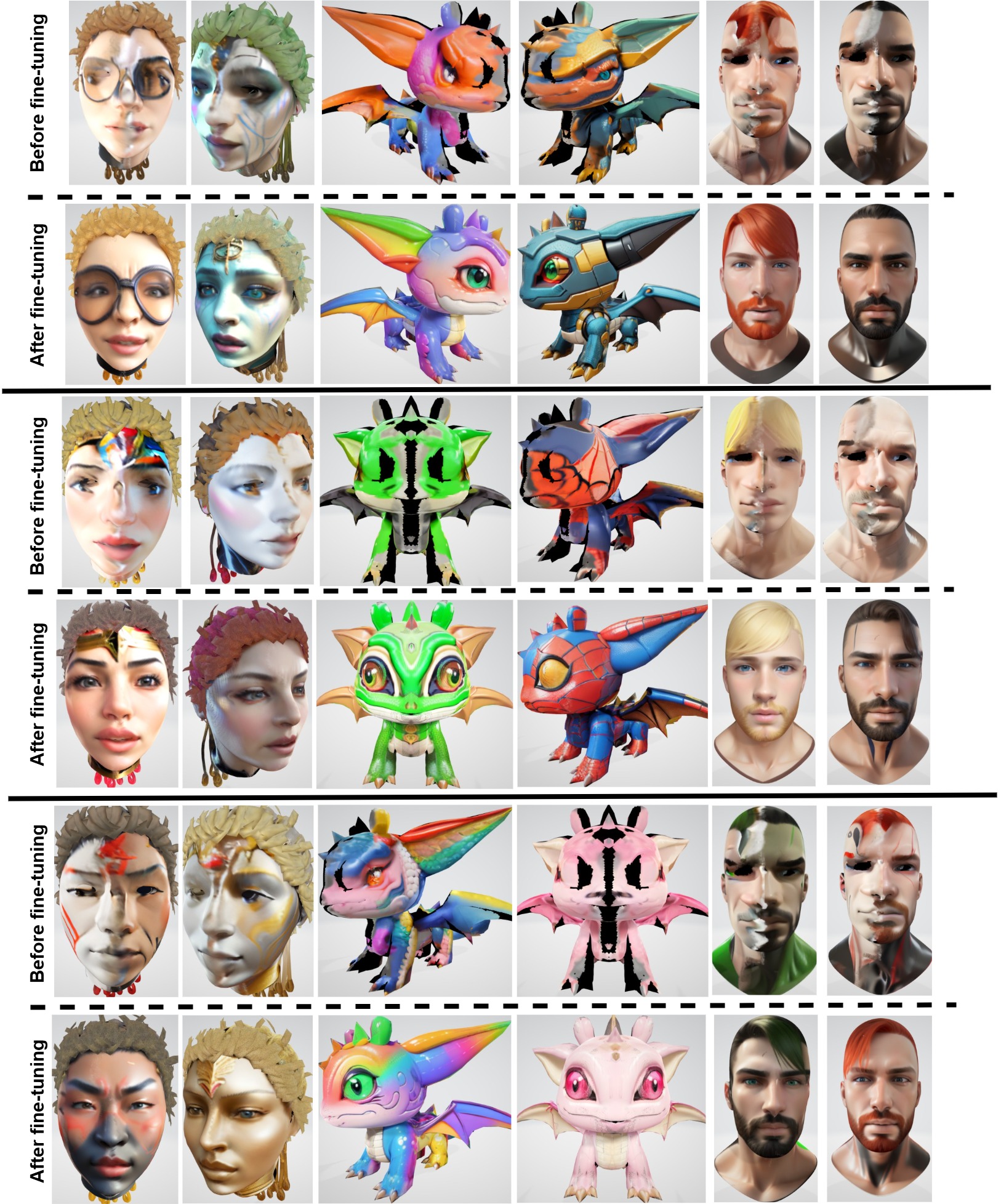}}
        \caption{Qualitative results of the camera-pose-aware texture generation experiment for different text prompts on different 3D mesh objects. The objective of this experiment is to learn optimal camera viewpoints such that, when the object is rendered and textured from these views, the resulting texture maximizes the average aesthetic reward. Consequently, by maximizing this reward, the model becomes invariant to the initial camera positions: regardless of where the cameras start, the training will adjust their azimuth and elevation to surround the 3D object in a way that yields high‑quality, aesthetically pleasing textures. We compare our fine‑tuning approach against InTeX \cite{InTex} and TEXTure \cite{TEXTure}. All methods share identical initial camera parameters; once initialized, each algorithm proceeds to paint the object. As shown, our method consistently outperforms both baselines, producing high‑fidelity, geometry‑aware textures across all shapes. This robustness stems from the aesthetic‑reward signal guiding camera adjustment during texture reward learning. In contrast, InTeX \cite{InTex} and TEXTure \cite{TEXTure} remain sensitive to poorly chosen initial viewpoints and thus often fail to generate coherent, high‑quality textures under suboptimal camera configurations.
        }
        \label{CameraPoseAwareTexGen_Qualitative}
        \end{center}
        \vskip -0.2in
    \end{figure*}

    \vskip -0.07in    
    \begin{figure*}[ht]
        \vskip 0.2in
        \begin{center}
        \centerline{\includegraphics[scale=0.255]{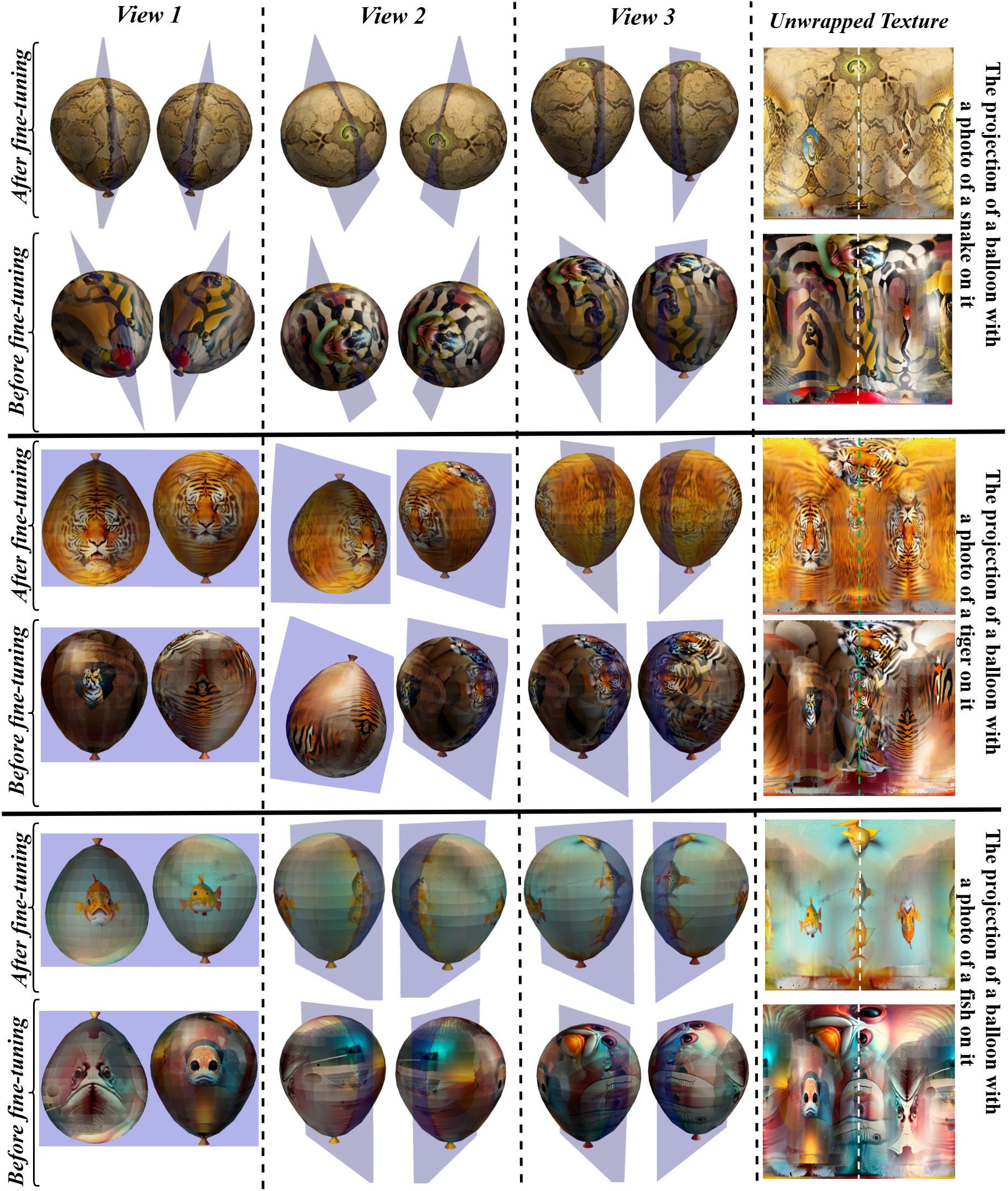}}
        \caption{Qualitative results of symmetry-aware experiment for different examples on a balloon mesh object. For each example (each rwo), we show the rendered 3D object from multiple viewpoints, alongside the corresponding texture images (rightmost column), which highlight the symmetric regions. A vertical dashed line marks the symmetry axis in each texture image. The purple plane passing through the center of the balloon in each viewpoint indicates the estimated symmetry plane of the object. As shown, compared to the pre-trained model \cite{InTex}, our method generates textures that are more consistent across symmetric parts of the mesh. Without symmetry supervision, patterns often differ noticeably between sides. In contrast, textures trained with the proposed symmetry reward exhibit visually coherent features across symmetric regions, demonstrating the reward’s effectiveness in enforcing symmetry consistency.}
        \label{Symmetry_Qualitative}
        \end{center}
        \vskip -0.2in
    \end{figure*}

    \vskip -0.07in    
    \begin{figure*}[ht]
        \vskip 0.2in
        \begin{center}
        \centerline{\includegraphics[scale=0.265]{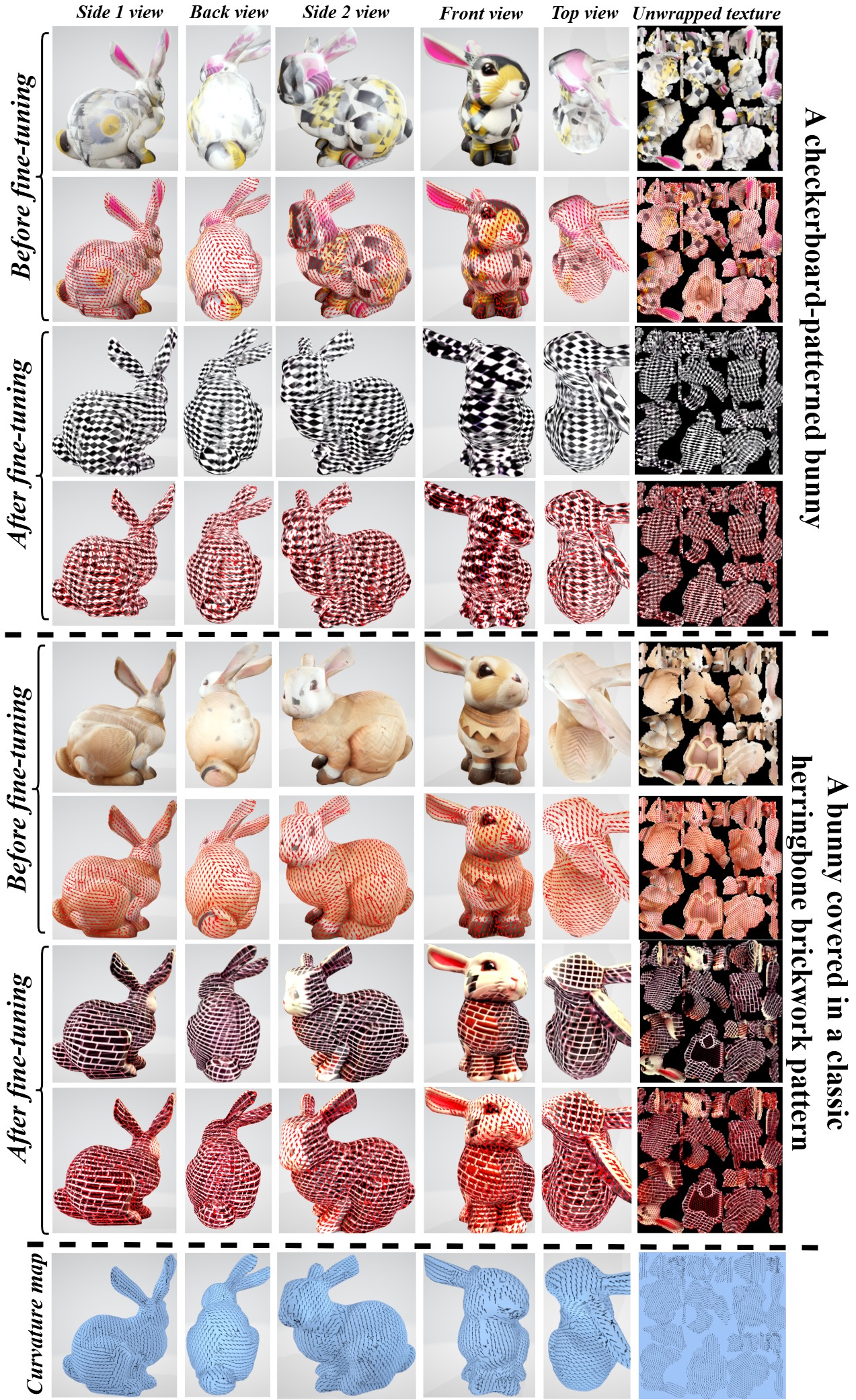}}
        \caption{Qualitative results of the geometry-texture alignment experiment on a rabbit (bunny) mesh. For each example, we show the rendered 3D object from multiple viewpoints, with the corresponding texture image in the rightmost column. Minimum curvature vectors, representing the underlying surface geometry, are visualized in the bottom row and overlaid on the textured objects for comparison. As shown, our method produces textures whose patterns align more closely with the mesh’s curvature directions, unlike InTex~\cite{InTex}. Moreover, a notable outcome in our results is the emergence of repetitive texture patterns after fine-tuning with the geometry-texture alignment reward. This behavior arises from the differentiable sampling strategy used during reward computation. Specifically, it encourages the model to place edge features at specific UV coordinates which ultimately results in structured and repeated patterns in the texture (see \cref{EmergenceRepetitivePatterns}).}
        \label{CurvTexAlign_Qualitative}
        \end{center}
        \vskip -0.2in
    \end{figure*}

    \vskip -0.07in    
    \begin{figure*}[ht]
        \vskip 0.2in
        \begin{center}
        \centerline{\includegraphics[scale=0.44]{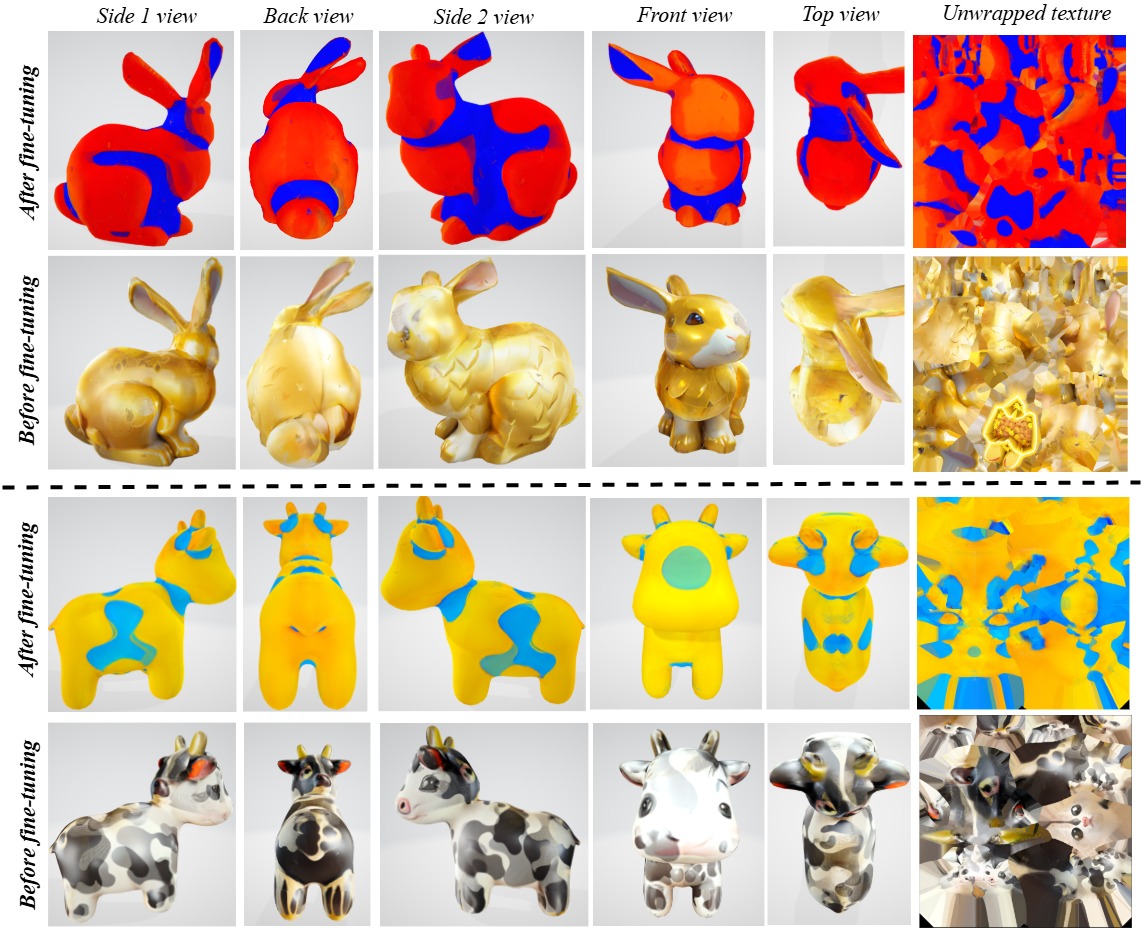}}
        \caption{Qualitative results of the geometry-guided texture colorization experiment on rabbit (bunny) a cow mesh objects. For each example, we show the rendered 3D object from multiple viewpoints, with the corresponding texture image in the rightmost column. The goal is to colorize textures based on surface bending intensity, represented by mean curvature, an average of the minimal and maximal curvature directions, on the 3D mesh. Specifically, the model is encouraged to apply warm colors (e.g., red, yellow) in high-curvature regions and cool colors (e.g., blue, green) in low-curvature areas. As illustrated, our method consistently adapts texture colors, regardless of initial patterns, according to local curvature and successfully maps warmth and coolness to geometric variation.}
        \label{GeometryGuidedTextureColorizationQualitativeResults}
        \end{center}
        \vskip -0.2in
    \end{figure*}

    \vskip -0.07in    
    \begin{figure*}[ht]
        \vskip 0.2in
        \begin{center}
        \centerline{\includegraphics[scale=0.44]{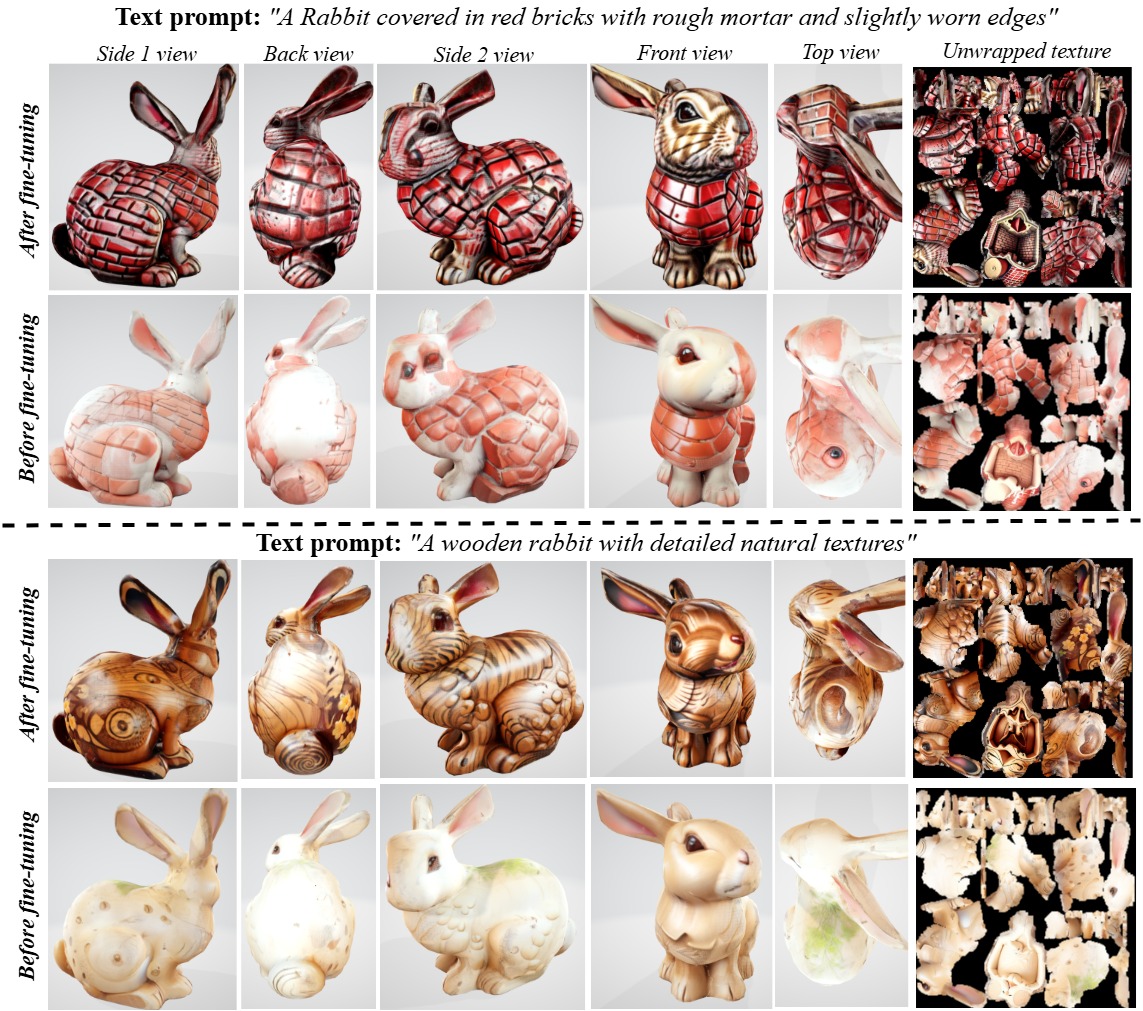}}
        \caption{Qualitative results of the texture features emphasis experiment on a rabbit (bunny) objects with different text prompts. For each example, we show the rendered 3D object from multiple viewpoints, with the corresponding texture image in the rightmost column.         This goal is to learn texture images with salient features (e.g., edges) emphasized at regions of high surface bending, represented by the magnitude of mean curvature. This encourages texture patterns that highlight 3D surface structure while preserving perceptual richness through color variation. As illustrated, our method enhances texture features, such as edges and mortar, in proportion to local curvature, a capability InTex \cite{InTex} lacks, often resulting in pattern-less (white) areas, particularly on the back and head of the rabbit.}
        \label{TextureFeatureEmphasis_Qualitative}
        \end{center}
        \vskip -0.2in
    \end{figure*}

    \begin{figure*}[ht]        
        \begin{center}
        \centerline{\includegraphics[scale=0.24]{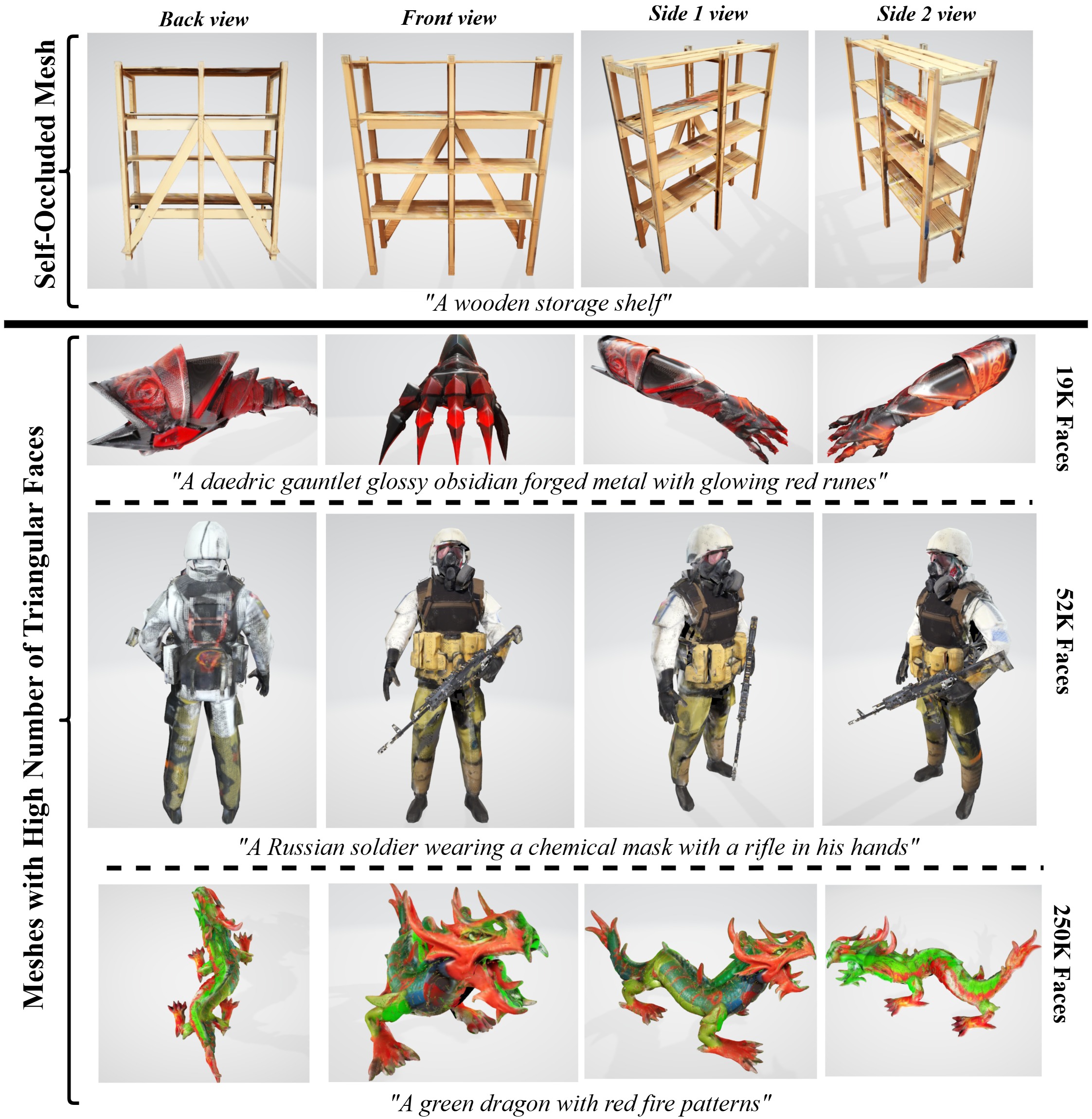}}
        \caption{Extended qualitative results for the camera-pose-aware reward. Top row: a self-occluded object; rows 2–4: complex, high-poly meshes. Because our method alters only texture (not geometry), self-occlusion and mesh resolution do not directly change the rendered outputs.}
        \label{ExtendedResults_ComplexObjects}
        \end{center}        
    \end{figure*}

    \vskip -0.07in    
    \begin{figure*}[ht]
        \vskip 0.2in
        \begin{center}
        \centerline{\includegraphics[scale=0.39]{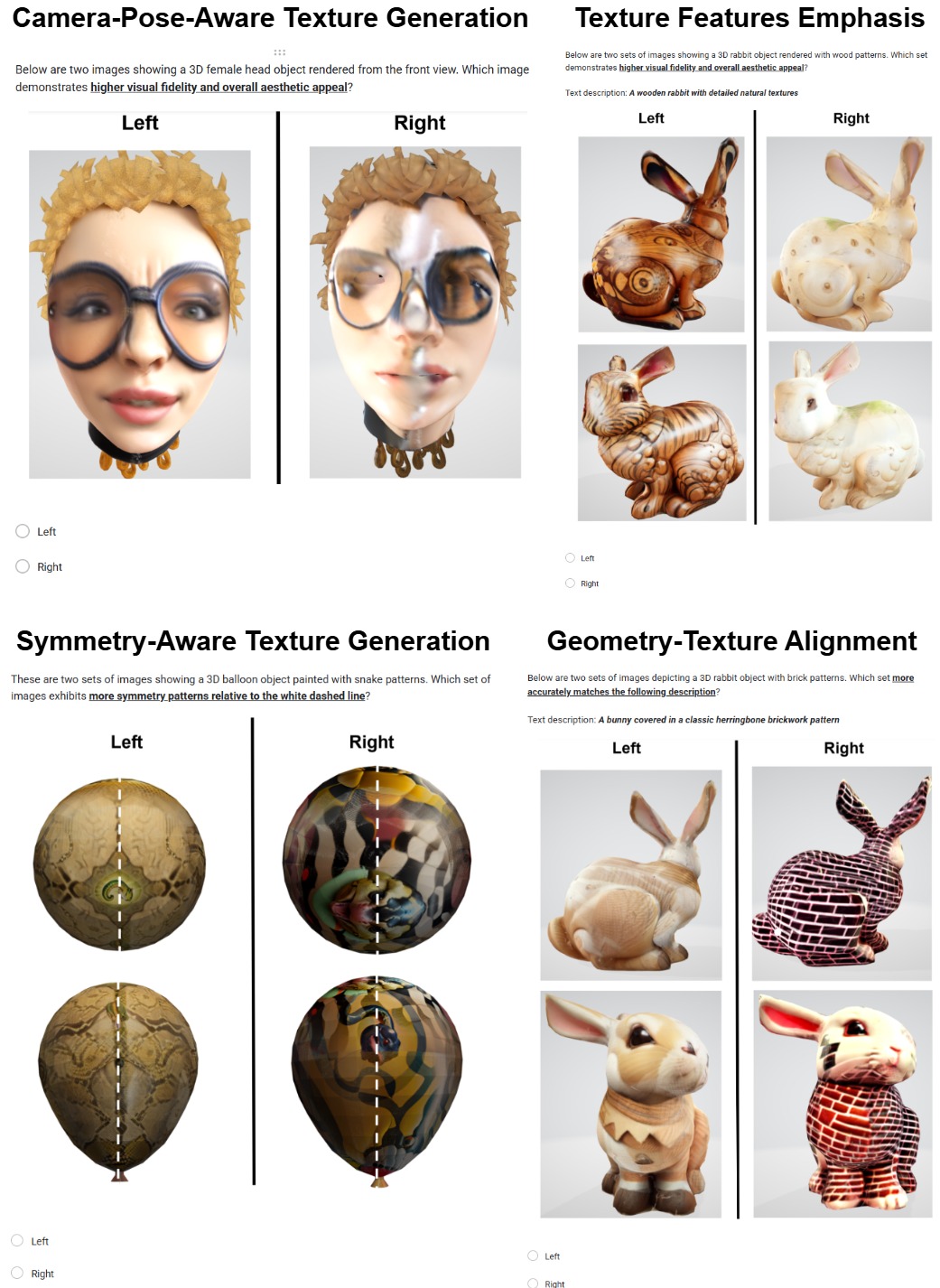}}
        \caption{Examples of four of the eleven questions in our user‐study questionnaire. We conducted a user‐preference study with 40 participants, each of whom completed 11 pairwise comparisons between textured 3D shapes generated by our method and by InTeX \cite{InTex}. Participants rated visual quality according to four criteria: fidelity, symmetry, overall appearance, and text‑and‑texture pattern alignment. Specifically, for the three reward variants, cam‐pose‐aware texture generation, symmetry‐aware texture generation, and geometry–texture alignment, we asked three questions apiece, and for the texture‐features‐emphasis reward we asked two questions. We then applied one‐sided binomial tests to each experiment to obtain p‑values (see bottom of \cref{QuantitativeResults_UserStudy}). As the table shows, the post‑reward‐learning results are significantly preferred over the pre‑learning outputs.}
        \label{UserStudyFigure}
        \end{center}
        \vskip -0.2in
    \end{figure*}

}

\vspace{25pt}

\section{Analysis of Failure Cases}
\label{FailureCases}
{  
    We observe two recurring failure modes under challenging inputs (see \cref{FailureCaseFigure}). First, the \textit{symmetry-aware} reward can introduce artifacts on non-extrinsically-symmetric or geometrically complex meshes. Our symmetry reward fits a dominant symmetry plane via Principal Component Analysis (PCA) on vertex coordinates, which implicitly assumes the vertex distribution is aligned with a clear axis/plane. When this assumption fails (e.g., complex geometry or uneven vertex distributions), PCA can return an incorrect plane and the symmetry reward then drives inconsistent, non-symmetric texture patterns (e.g., in the top two rows of Fig.~\ref{FailureCaseFigure}, the missing eye on one side of the rabbit’s face while present on the other, or asymmetric texture patterns on the face of the 3D female bust). Second, the \textit{camera-pose-aware} reward, designed to remove manual viewpoint tuning by learning camera azimuth/elevation, can fail if camera initialization is extremely poor (e.g., cameras placed too far from the object or at extreme off-angles). In these cases, the learned viewpoints may not cover the full surface and parts of the mesh remain untextured (black regions), as shown in Fig.~\ref{FailureCaseFigure} (bottom row).

    \begin{figure*}[ht]        
        \begin{center}
        \centerline{\includegraphics[scale=0.47]{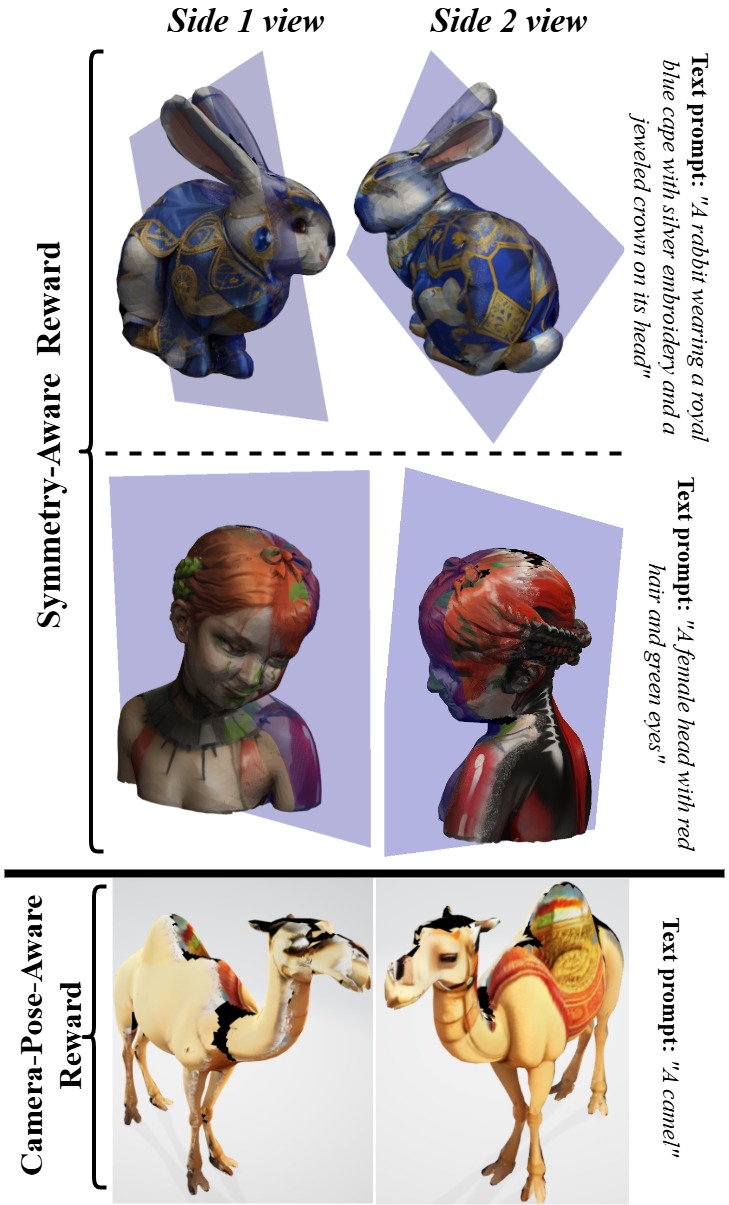}}
        \caption{Failure cases. \textbf{Top}: artifacts introduced by the symmetry-aware reward on non-extrinsically symmetric (rabbit and bust). Our PCA-based symmetry reward can produce an incorrect symmetry plane when vertex distributions are uneven, causing the symmetry objective to drive inconsistent or non-symmetric textures (e.g., the missing eye on one side of the rabbit’s face while present on the other, or asymmetric texture patterns on the face of the 3D female bust). \textbf{Bottom}: failure of the camera-pose-aware reward on the Camel. Extremely poor camera initialization (e.g., camera placed too far from the object or at extreme off-angles) can prevent learned viewpoints from covering the full surface, leaving untextured (black) regions.}
        \label{FailureCaseFigure}
        \end{center}        
    \end{figure*}
}

\end{document}